\documentclass[preprint,5p]{elsarticle}

\usepackage{amssymb}
\usepackage{amsmath}
\usepackage{graphicx}%
\usepackage{multirow}%
\usepackage{amsmath,amssymb,amsfonts}%
\usepackage{amsthm}%
\usepackage{mathrsfs}%
\usepackage[title]{appendix}%
\usepackage{xcolor}%
\usepackage{textcomp}%
\usepackage{manyfoot}%
\usepackage{booktabs}%
\usepackage{algorithm}%
\usepackage{algorithmicx}%
\usepackage[noend]{algpseudocode}%
\usepackage{listings}%
\usepackage{pifont}
\newcommand{\cmark}{\ding{51}}%
\newcommand{\xmark}{\ding{55}}%
\usepackage{url}

\journal{Robotics and Autonomous Systems}

\begin{document}

\begin{frontmatter}

%% Title, authors and addresses

%% use the tnoteref command within \title for footnotes;
%% use the tnotetext command for theassociated footnote;
%% use the fnref command within \author or \affiliation for footnotes;
%% use the fntext command for theassociated footnote;
%% use the corref command within \author for corresponding author footnotes;
%% use the cortext command for theassociated footnote;
%% use the ead command for the email address,
%% and the form \ead[url] for the home page:
%% \title{Title\tnoteref{label1}}
%% \tnotetext[label1]{}
%% \author{Name\corref{cor1}\fnref{label2}}
%% \ead{email address}
%% \ead[url]{home page}
%% \fntext[label2]{}
%% \cortext[cor1]{}
%% \affiliation{organization={},
%%             addressline={},
%%             city={},
%%             postcode={},
%%             state={},
%%             country={}}
%% \fntext[label3]{}

\title{Multimodal Human-Intent Modeling for Contextual Robot-to-Human Handovers of Arbitrary Objects}
%\tnotetext[t1]{This work was supported by the National Science Foundation (NSF) under award no. 2204528.}

%% use optional labels to link authors explicitly to addresses:
%% \author[label1,label2]{}
%% \affiliation[label1]{organization={},
%%             addressline={},
%%             city={},
%%             postcode={},
%%             state={},
%%             country={}}
%%
%% \affiliation[label2]{organization={},
%%             addressline={},
%%             city={},
%%             postcode={},
%%             state={},
%%             country={}}

\author[1]{Lucas Chen\fnref{fn1}}
%\ead{chen4007@purdue.edu}

\author[1]{Guna Avula \fnref{fn1}}
%\ead{gavula@purdue.edu}

\author[1]{Hanwen Ren\corref{cor1}}
\ead{ren221@purdue.edu}

\author[1]{Zixing Wang}
%\ead{wang5389@purdue.edu}

\author[1]{Ahmed H. Qureshi}
%\ead{ahqureshi@purdue.edu}

\cortext[cor1]{Corresponding author}

\fntext[fn1]{Indicates equal contribution}

%\author{Lucas Chen, Guna Avula, Hanwen Ren, Zixing Wang, Ahmed H. Qureshi} %% Author name

%% Author affiliation
\affiliation[1]{organization={Departmant of Computer Science, Purdue University},%Department and Organization
            addressline={305 N University St}, 
            city={West Lafayette},
            state={IN 47907},
            %postcode={47907}, 
            country={USA}}

%% Abstract
\begin{abstract}
%% Text of abstract
Human-robot object handover is a crucial element for assistive robots that aim to help people in their daily lives, including elderly care, hospitals, and factory floors. The existing approaches to solving these tasks rely on pre-selected target objects and do not contextualize human implicit and explicit preferences for handover, limiting natural and smooth interaction between humans and robots. These preferences can be related to the target object selection from the cluttered environment and to the way the robot should grasp the selected object to facilitate desirable human grasping during handovers.  Therefore, this paper presents a unified approach that selects target distant objects using human verbal and non-verbal commands and performs the handover operation by contextualizing human implicit and explicit preferences to generate robot grasps and compliant handover motion sequences. We evaluate our integrated framework and its components through real-world experiments and user studies with arbitrary daily-life objects. The results of these evaluations demonstrate the effectiveness of our proposed pipeline in handling object handover tasks by understanding human preferences. Our demonstration videos can be found at https://youtu.be/6z27B2INl-s.
\end{abstract}

%%Graphical abstract
%\begin{graphicalabstract}
%\includegraphics{grabs}
%\end{graphicalabstract}

%%Research highlights
%\begin{highlights}
%\item Proposal of a novel multimodel approach for object handover.
%\item Robot grasp generation that considers human implicit and explicit preferences.
%\item Unified system that maps human instructions to robot actions.
%\item User study demonstrates the effectiveness and robustness of the proposed approach. 
%\end{highlights}

%% Keywords
\begin{keyword}
%% keywords here, in the form: keyword \sep keyword
Physical Human-Robot Interaction \sep Grasping \sep Human-Centered Robotics \sep Human-Robot Object Handover
%% PACS codes here, in the form: \PACS code \sep code

%% MSC codes here, in the form: \MSC code \sep code
%% or \MSC[2008] code \sep code (2000 is the default)

\end{keyword}

\end{frontmatter}

%% Add \usepackage{lineno} before \begin{document} and uncomment 
%% following line to enable line numbers
%% \linenumbers

%% main text
%%

%% Use \section commands to start a section

\section{Introduction}
Assistive robots need to be able to hand over unknown objects to and from humans in different environments from remote places to reduce their human partners' workload \cite{ortenzi2022object, DUAN2024100145}. For example, in hospitals, they would need to hand over items like medicines, thermometers, and food trays to different patients to assist healthcare workers. Similarly, in elderly care, robots would be expected to assist with tasks such as fetching and handing over items like TV remotes, mobile phones, keys, utensils, and books. In factory settings, robots with these abilities can significantly improve work efficiency by handing over or fetching various tools for their collaborating workers.

Despite the widespread applications and the need for solving object handover tasks, the problem of seamlessly selecting and handing over arbitrary objects from remote places remains relatively unexplored. We believe the traits of an ideal object handover system would be the following:
\begin{itemize}
    \item A way for humans to naturally select the target object in remote places for handover using verbal and non-verbal cues. 
    \item A mechanism to contextualize human implicit or explicit preferences of simultaneously grasping an object with the robot.
    \item A robotic grasping methodology that is capable of handling arbitrary unknown objects, adheres to human preferences, and produces an appropriate grasp for object handover.
    \item A compliant robot handover motion generator that avoids collisions to seamlessly execute the handover procedure.  
\end{itemize}

The existing approaches to solving object handover lack one or more of the ideal features mentioned above~\cite{Aleotti2014AnAS, Dang2012SemanticGP, 5326233, liu2024hoi4d, wei2024grasp, Kim1992HandoverOA}. Firstly, the existing work mostly considers pre-selected target objects for handover. However, in practice, humans would need to specify the target objects for the robot to retrieve and handover. Furthermore, these specifications of target objects are usually for distant objects that are not directly within human reach. Secondly, most methods make one or more of the following unrealistic assumptions, such as the availability of exact 3D object models and predefined robot grasps, and often do not contextualize human preferences for solving handover tasks. An exception includes our prior work called CoGrasp \cite{10160623}, which contextualizes human preferences into generating robot grasps for arbitrary unknown objects. 

However, CoGrasp only considers human implicit social preferences, which means it leverages data-driven models to determine how a human would prefer to hold an object simultaneously with the robot. It does not account for explicit preferences that a human may specify on the fly during collaboration. Furthermore, CoGrasp also assumes pre-selected target objects for handover. Similar to CoGrasp, most prior methods \cite{8206205, 8341961} assume pre-selected objects for handover. They overlook remote scenarios in which the human may be far from the objects and must instruct the robot to bring a specific item. To the best of our knowledge, no unified approach exists that enables robots to seamlessly select and hand over target objects from remote locations using commands only in vision-language space without requiring any 3D models of objects. Therefore, in this paper, we aim for an approach that maps vision and language-based human commands to robot action space for solving object grasping and handover tasks from distant places.

The specification of the target, distant object requires modeling of human intention to interpret verbal and non-verbal cues. Most existing approaches rely on single-modal signals to infer human intention. For example, some studies investigate the use of electroencephalography (EEG) for controlling various digital interfaces, such as a cursor~\cite{eeg_cursor} or a robot arm~\cite{eeg_robot_arm}. However, signals from EEG are generally noisy and lack precision for our task. Recent advances in natural language processing and computer vision have brought significant progress in this field. Through their understanding of free-form language, language models such as BERT~\cite{devlin2019bert}, T5~\cite{raffel2023exploring}, ELECTRA~\cite{clark2020electra}, GPT~\cite{brown2020language} have been critical in building new models integrating language with 2D images~\cite{LISA} and 3D scenes~\cite{3d_llm_segmentation_1,3d_llm_segmentation_2,3d_llm_segmentation_3}. While understanding spatial information within a sentence reduces ambiguity, users providing detailed descriptions is not always guaranteed. This is complemented by gaze detection, which tracks eye gaze and posture to distinguish similar objects without semantic descriptions. While gaze often requires extensive hardware~\cite{gaze_survey}, recent research such as MPIIFaceGaze \cite{zhang2017mpiigaze} developed a robust approach to derive gaze direction by combining both eye tracking and facial orientation from a monocular image~\cite{zhang19_pami,ccakir2023reviewing}. 

Therefore, in our framework, we leverage natural languages and gaze to demystify human commands in vision-language space and infer their intentions to select the remotely placed target object. Once the object is selected, our framework performs the object handover operation by contextualizing human implicit and explicit preferences to generate robot grasps and compliant handover motion sequences. In summary, the key contributions of our proposed approach are as follows:
\begin{itemize}
    \item A multimodal approach for target object selection that demystifies human verbal and non-verbal cues using vision, language, and gaze.
    \item A contextual robot grasp generator for arbitrary unknown objects that extends our prior work CoGrasp \cite{10160623} to follow human implicit and explicit preferences when simultaneously grasping the object with the robot during handover. 
    \item A compliant robot motion generator that outputs smooth trajectories for realizing robot-to-human object handover operation.
    \item An end-to-end unified approach that maps human verbal and non-verbal commands in vision-language space to the robot's action space for selecting, grasping, and handing over the desired object to humans.
\end{itemize}
We evaluate our individual components as well as the integrated pipeline through a series of real-world experiments and user studies. Our results demonstrate that the proposed pipeline seamlessly solves the object handover tasks by contextualizing human implicit and explicit preferences.

\section{Related Work}\label{sec2}
In this section, we present the related work focused on modeling human verbal and non-verbal cues and addressing human-to-robot object handover tasks.

\subsection{Human Verbal and Non-verbal Behavior Modeling}\label{sec1sub1}
In addition to the kinematic task of physically passing objects to the human, another challenge lies in modeling human intent to deduce the desired outcome of the handover. The work~\cite{Strabala2013TowardSH} highlights that natural human-human handovers involve intricate signaling of body language, social awareness, and gaze to facilitate smooth and natural transitions. Here, we review the different signals used in human-robot handover to better understand human intent.  

First, we describe means of referring to objects via text and discuss their limitations. This is often encapsulated in the referring segmentation task ~\cite{qiao2020referring}, where language prompts are used to query for objects in an image or 3D representation. For example, in~\cite{Langer2022ILG,zhang2024invigorate}, a vision-language model is used to translate the user's requests into object selections. Although models vary in the scene representation, such as images \cite{chen2022gscorecam, LISA, lueddecke22_cvpr}, meshes~\cite{Decatur20223DHL}, or point clouds~\cite{song2023learning, Zhu2022PointCLIPVP, Ngyen2023OpenVocabularyAD}, they share the text input modality. However, \cite{zhang2024invigorate} indicates that text is often ambiguous and lacks precision in isolating semantically similar objects and aims to solve this via an interactive question-answering process. In work similar to ours, \cite{Tang2023TaskOrientedGP} regresses object grasping poses from a vision-language model but highlights language grounding is a significant bottleneck to accuracy. To tackle this inherent ambiguity in language, \cite{Ding2020PhraseClickTA} propose a joint method that combines cursor pointing with language to improve referring segmentation accuracy. 

Another line of work uses human pose estimation to assist robots in interpreting human intentions \cite{Leal2020INITIATINGOH}. These methods often leverage deep pose prediction methods \cite{9515402, choi2022preemptive}. For example, pointing gestures serve as clues for specifying objects and directions \cite{Cosgun2015DidYM}. Moreover,~\cite{trick2019multimodal} demonstrates that pose is too uncertain to function as a lone source of human intent. Furthermore, in the handover task, the objects are usually far away and unreachable by humans. Therefore, pointing actions through gestures won't help resolve the ambiguity in the language commands.

To overcome the need for a human to move their body, such as in moving digital cursors or changing body posture to indicate their intention, methods exist that leverage Electroencephalograms (EEGs) to infer human intentions. These approaches use EEGs to identify preparatory neural activities associated with motor tasks, which could inform more responsive and adaptive robot behaviors \cite{rajabi2023detecting, sharma2022towards}. However, EEGs are characterized by a high signal-to-noise ratio and require operators to take forceful actions to properly recover a pattern \cite{cooper2020eeg}. 

Gaze has been proposed to augment the pitfalls of language in referral. While language encapsulates semantic properties, or ``what'' object is desired, the gaze is often used to communicate the spatial context, or ``where'' the object is \cite{Vasudevan2018ObjectRI}. Moreover, the gaze is produced passively and directly reveals human intent and attention during object manipulation \cite{Tian2024GazeguidedHI}. This makes it an attractive modality for understanding the user's intent during handover. Studies indicate that gaze behavior can provide users with non-verbal cues about the robot's intentions and understanding \cite{staudte2008utility, Moon2014MeetMW}, improving an interaction's naturalness and efficiency \cite{understandtranfer, Aronson2018EyeHandBI}. 

While gaze tracking has been a popular task for object selection ~\cite{dwellpaulus} in indoor \cite{eyepointing} and outdoor\cite{shao2015eyelasso} scenarios, classical gaze tracking comes with the challenges of specialized hardware, high cost and unwieldy form factors. The rise of deep learning methods \cite{krafka2016eye} has made gaze estimation possible using conventional monocular cameras. Although older methods \cite{krafka2016eye, Deng2017MonocularF3} relied on accurate eye segmentation, recent methods \cite{zhang19_pami} directly regress gaze from head images, even with obscured eye \cite{FischerECCV2018}, and in dynamically lit outdoors \cite{kellnhofer2019gaze360}. This is expedited by the rise of gaze datasets~\cite{Smith2013GazeLP,McMurrough2012AnET,Mora2014EYEDIAPAD,4449972} that allows learning methods to better track extreme head poses under a large variety of lighting conditions and environments. These models exhibit superior performance to the older geometric approaches that rely on unobstructed images of the eyeball~\cite{Hennessey2006ASC}. In this work, we leverage a recent high-performing gaze tracking model to produce accurate spatial signals. Our experiment results show that gaze helps eliminate verbally and visually similar object candidates from the true target object.

\subsection{Human-Robot Object Handover}\label{sec2sub2}

Human-robot handover can be categorized as Human-to-robot (h2r) and Robot-to-human (r2h) handover tasks. While there exist various methods~\cite{Pan2017AutomatedDO, Micelli2011PerceptionAC, 8579107, 6281385, 9341004, yang2021reactive, 6343845, 4415256,Christen2023SynH2RSH} that address the h2r task, our focus is on solving the r2h handover tasks. The r2h handover tasks are challenging as the robot needs to infer human social preferences when handing over their desired objects. These tasks can be divided into two subproblems, namely contextual robot grasp generation and r2h handover motion generation. 

Contextual grasping involves generating robot grasps with context about underlying tasks. Most of the prior work in this domain does not consider human-in-the-loop settings~\cite{Antanas2018SemanticAG, Aleotti2014AnAS, Detry2012GeneralizingGA, Dang2012SemanticGP, Song2010LearningTC, liu2020cage}. Their prime focus is generating contextual grasps for daily life tasks such as pouring, cutting, lifting, etc. These tasks require the robot to hold the object in a certain way to accomplish the underlying task. Recently, large language models have also been employed to select an appropriate grasp for the underlying tasks~\cite{zheng2024gaussiangrasper,wei2024grasp}. Some of these methods also consider the object handover tasks~\cite{Kim1992HandoverOA}. However, they do not explicitly model human social preferences and instead focus on leaving enough portions of the object for humans to simultaneously grasp the object. Human preference often needs robots to hold objects in a specific way. For instance, when handing over sharp objects, the human would prefer to hold them by the handle; therefore, the robot should grasp them by the blade part. Such social preferences are not contextualized in the grasping methods mentioned above. An exception includes our prior work known as CoGrasp~\cite{10160623}, which infers human preferences using data-driven models for human hand grasp generation and leverages them to select non-conflicting robot grasps. However, CoGrasp only implicitly models human preferences through data-driven models and does not consider preferences given explicitly by humans. In this work, we also extend CoGrasp to tackle both implicit and explicit human preferences and integrate the resulting model into our unified object handover framework.

The r2h handover motion generation primarily requires the robot to adhere to collision avoidance and safety constraints. The existing work~\cite{Mainprice2010PlanningSA,Sisbot2010SynthesizingRM,Micelli2011PerceptionAC, 5326233, Kajikawa2000TrajectoryPF} systemically consider human factors to address safety and collision-avoidance. For instance, in~\cite{Sisbot2010SynthesizingRM}, the proposed method addresses safety constraints by a soft motion generator conditioned on human vision range and bio-mechanical constraints to avoid unsafe contact. Likewise, the method in~\cite{Micelli2011PerceptionAC} learns robot motion constrained by the collision-free social norm and the uncertainties by imitating human social behavior. Similarly,~\cite{5326233} replaces the traditional trapezoidal velocity profile with a bio-inspired trajectory generator to reduce jerks for less aggressive behaviors. Our proposed framework also aims toward generating handover motion without jerks by optimizing the robot velocity profile. To achieve this objective, we leverage the Riemannian Motion Policy (RMP)~\cite{Ratliff2018RiemannianMP} framework that has been demonstrated to generate fluid motion without jerks and scales to higher dimensional robot configuration spaces.

\section{Methodology}\label{sec3}
\begin{figure*}[ht]
\centering
\includegraphics[trim = {0.7cm 1.3cm 1cm 2cm}, clip, width=1.0\textwidth]{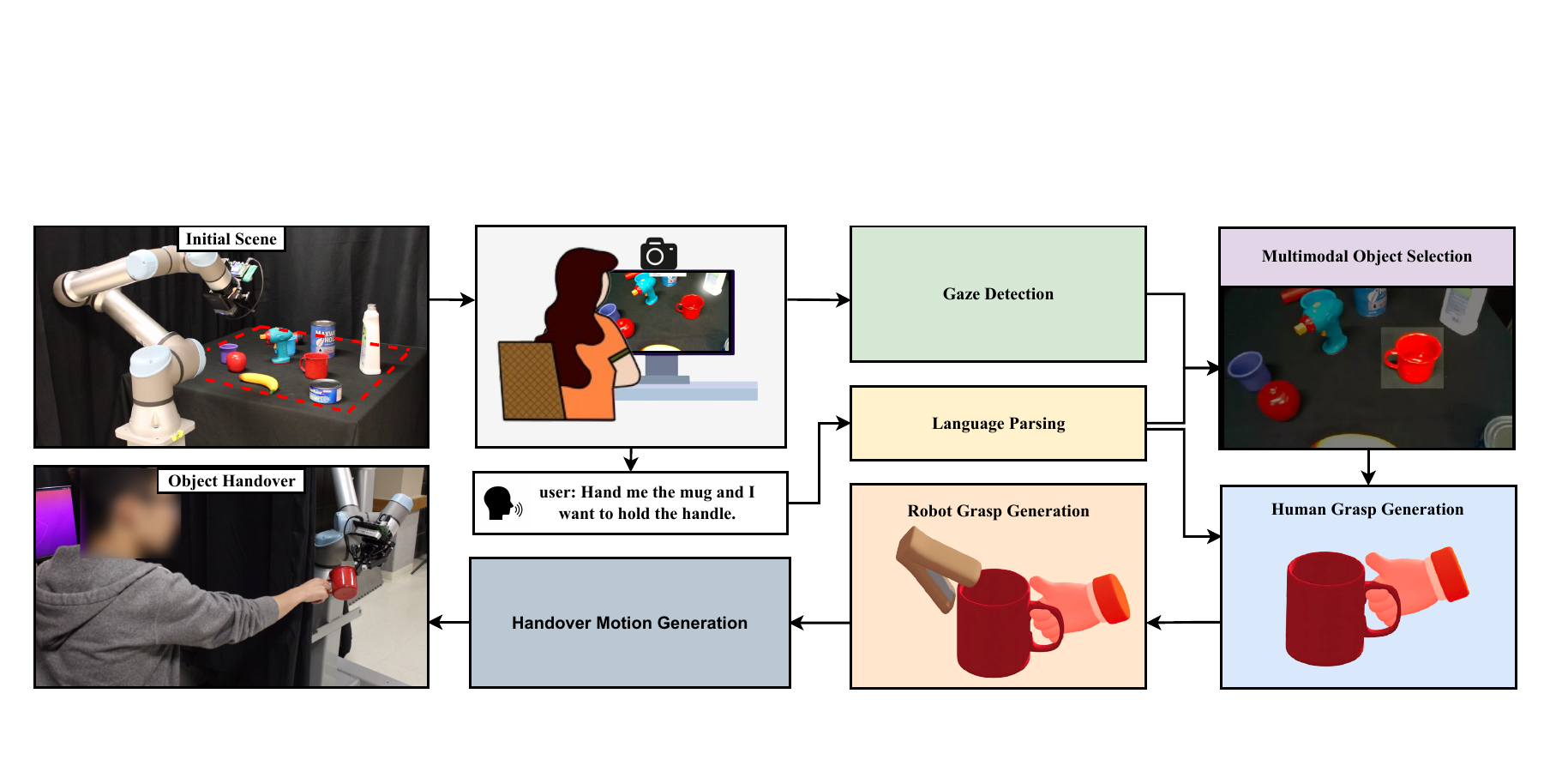}
\caption{Human Intention-Aware Object Handover System: The robot captures the RGB-D image of the scene through its in-hand camera and sends it to the user's monitor. The user then identifies an object of interest by giving out clues in the form of gaze and natural language prompts. Based on the pre-processed user's inputs, the multimodal object selection module figures out the desired object and segments it out from the image. In order to enforce safe and smooth human-robot interactions, our approach generates the most suitable robot grasp by considering the user's grasp preferences. Finally, the handover motion generation module moves the robot to grasp the detected object of interest and eventually hand it over to the user.}
\label{pipeline}
\end{figure*}

This section formally reveals the key components of our precise multimodal object handover approach, namely gaze detection, language parsing, multimodal object selection, robot grasp generation, and handover motion generation. The overall structure of our method is shown in Fig.~\ref{pipeline}.

%\vspace{-30pt}

\subsection{Problem Definition}\label{sec3sub1}
Let a robot arm $R$ with an in-hand camera placed beside a tabletop environment $S$ filled with objects $\{O_1,\dots,O_n\}$. A user $U$ who does not have direct access to the environment sits in front of a monitor with a facial camera attached. To enable an intuitive and informative human-robot interaction, our system captures both the color $I_c$ and depth map $I_d$ of the scene $S$ with potential objects for grasping. The color information is further projected to the user's monitor. Once the scene is observed, the user is allowed to use both gaze as well as language prompts to identify an object of interest $O_u$ for the robot to grasp and handover. As a result, our system records the spoken request transcription $T$ from the microphone as well as an image stream $\{I_f\}_n$ from the facial camera of size $n$. We will use notation $\{A\}_B$ in the rest of this paper to represent a set of size $B$ that is composed of element $A$. Once the user's inputs are captured, our goal is summarized in order as follows:
\begin{enumerate}
    \item Identify the user's object of interest $O_u$ and the corresponding grasping preference using the recorded language transcription $T$ and facial camera image stream $\{I_f\}_n$. 
    \item Synthesize a stable user-aware robot grasping position $G_{O_u}$ for the desired object that aligns with the user’s spoken requirements.
    \item Plan a collision-free path for the robot to grasp the object of interest $O_u$ and hand it over safely to the user. 
\end{enumerate}

\subsection{Gaze Detection}\label{sec3sub2}
The Gaze detection module captures the user's eye focal point on the monitor through a stream of facial camera images. The logic of designing this subsystem lies in the fact that existing literature shows that human eye gaze contains strong communicative signals \cite{nakano2010estimating}. In our case, the user's eyesight is more likely to focus on the desired object than others while giving the language instructions. Thus, analyzing the facial camera images would give us a clue about the locations of the user's eye focal points, which as a result lead to the desired object. This module is extremely useful in solving the ambiguity when multiple copies of the same object are presented in the scene.

The input to this subsystem is the image stream $\{I_f\}_n$ captured by the facial camera mounted on top of the user's monitor. To accurately acquire the user's eye focal point, we evaluated several existing models for our setup. For example, Gaze360 \cite{kellnhofer2019gaze360}, trained on unconstrained outdoor datasets, performs well with large head movements but not in our monitor-based scenario. Instead, we select MPIIFaceGaze \cite{zhang2017mpiigaze}, trained on people looking at mobile devices, which is more precise for subtle eye movements but doesn't support extreme head pose variations. Despite this, the head direction still matters in our setup, which may feature a greater downward average direction compared to the direction in the MPIIFaceGaze dataset. Therefore, we propose an ensemble model combining the head direction $ V_h \in \mathbb{R}^3$ from MediaPipe \cite{lugaresi2019mediapipe} and the gaze direction $V_g \in \mathbb{R}^3$ from MPIIFaceGaze as the ultimate user's gaze direction $V_u \in \mathbb{R}^3$ for increased robustness as follows: 
\begin{equation}
    V_u = \alpha V_h + (1 - \alpha) V_g
\end{equation}
where $\alpha \in [0, 1]$ is a weighting factor and $V_u$ is normalized to be a unit vector.

To project the corrected gaze direction $V_u$ to screen coordinates, we describe the monitor as a plane $P_m \in \mathbb{R}^2$ whose normal is defined by the cross product of two orthogonal vectors along its width and height. Then, assuming the bottom left corner of the monitor locates at $B_m \in \mathbb{R}^3$ in the global coordinate system and the two orthogonal vectors are represented as $(V_{m_1} \in \mathbb{R}^3 , V_{m_2} \in \mathbb{R}^3)$, the location of all pixels at $(\lambda, \mu) \in \mathbb{R}^2$ inside the monitor frame can be represented as $B_p \in \mathbb{R}^3$ in the global coordinate system\\
\begin{equation}\label{eq2}
    B_p = B_m + \lambda V_{m_1} + \mu V_{m_2}
\end{equation}
On the other hand, the user's gaze can be defined by the head location $B_u \in \mathbb{R}^3$ and the pre-calculated gaze direction $V_u$ shown in Equation (1). The head location $B_u$ is calibrated for each participant by measuring the position of their eyes when sitting upright and facing the monitor. Thus, the final gaze vector can be written as $B_g \in \mathbb{R}^3$ in the global coordinate system 
\iffalse
gaze comes from the user whose head location is at $B_u \in \mathbb{R}^3$, the ending point of it {\color{red}can be written as $B_g\in \mathbb{R}^3$ in the global coordinate system, where}\\
\fi
\begin{equation}\label{eq3}
    B_g = B_u + \sigma V_u
\end{equation}
where $\sigma$ denotes the travelling distance of the user's gaze in direction $V_u$.

Using Equation \ref{eq2} and \ref{eq3}, we can obtain the user's pixel of interest $G_{k} = (\lambda, \mu)$ for each image in the stream $I_k \in \{I_f\}_n$ by solving $B_p = B_g$ for $(\lambda, \mu, \sigma)$ indicating the intersection point between user's gaze and the monitor plane. Note that Equations \ref{eq2} and \ref{eq3} are in 3D space. Therefore, expanding them into their x, y, and z components gives us three separate scalar equations with three unknowns $(\lambda, \mu, \sigma)$. Hence, this system of equations is fully determined and can be solved directly, providing a unique intersection point between the user’s gaze and the monitor plane.

\begin{figure}[htbp]
\centering
\includegraphics[trim = {3cm 4.5cm 6cm 3cm}, clip, width = 9 cm]{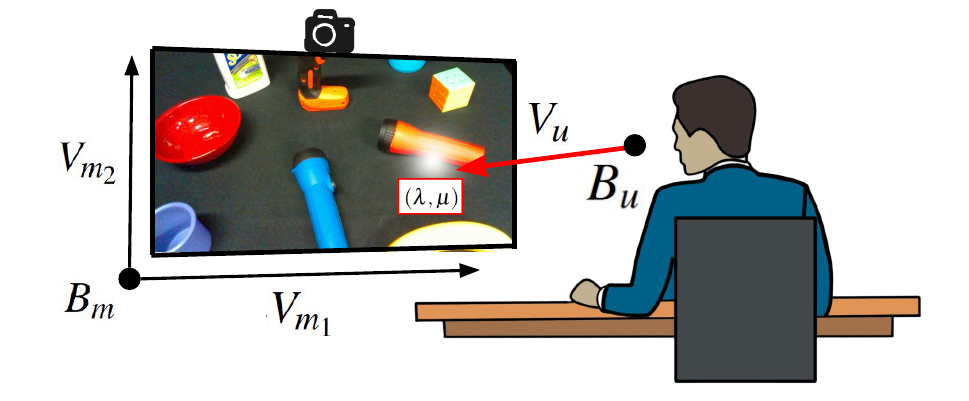}
\caption{This figure depicts the essential points and vectors used in our gaze detection approach. By computing the intersection point between the user's gaze vector and the monitor plane, our method accurately obtains the user's pixel of interest $G_k = (\lambda, \mu)$ in the image, represented as the bright dot.}

%{of our gaze setup, with key vectors and points labeled. By knowing the pose of the monitor, user's head, and gaze direction, we can find the coordinates of the pixel of interest $G_k$. We further generate the heatmap $H_g$ from $G_k$}.}
\label{fig:gaze_math}
%\vspace{-0.2in}
\end{figure}

A graphical representation of our gaze method is presented in Fig. \ref{fig:gaze_math}. Next, we apply an exponential moving average (EMA) technique to all the calculated pixels of interest in the image stream autoregressively to ensure a more stable and accurate gaze detection:
\begin{equation}
    G_{k+1}' = \beta G_{k+1} + (1-\beta)G'_{k}
\end{equation}
where $k \in [1, n-1]$ and $\beta$ is a smoothing factor. 

Finally, we generate a discrete heatmap $H_g$ of the same size as the scene image $I_c$ by creating a 2D Gaussian distribution centered at the resulting pixel of interest $G_n'$. The standard deviation of the distribution on each axis is set to 57 pixels, while the covariance between the axes is 0. We determine these values via cross-validation and qualitative analysis of our gaze system performance. The Gaussian distribution assigns a smooth non-zero probability density value to each pixel in $H_g$, indicating its chance of being the actual user's eye focal point. This design aims to resolve the cases where the inaccuracies in the user's gaze cause the pixel of interest to be located on the wrong object, especially in cluttered environments. Compared with assigning a single value to one single pixel in the image, the introduction of the Gaussian distribution allows the pixels of the desired object to always have decent gaze probabilities, leading to higher chance of successful object detection and handover.  

\subsection{Language Parsing}\label{sec3sub3}
Our language parsing module isolates the object of interest from a natural language command given by the user. Moreover, it deduces whether a certain part should be held by the robot or the user using contextual information. This module serves to enhance the performance of object selection by parsing the object name and the holding pose preference.

Given the user instruction transcription $T$, the first step is to figure out the user's desired object $T_O$ in the text form. Although manual parsing performs well for a selected set of test sentences, the users tend to produce a great diversity of sentences that could not all be accounted for without learning-based approaches. Thus, in order to accomplish this parsing task on a diverse set of language commands, we opt for a roBERTa model ~\cite{liu2019roberta} trained on the Stanford SQUAD dataset, which is proved to provide more consistent and accurate results than traditional language parsing techniques. Utilizing the pre-trained roBERTa model, the user's desired object $T_O$ can be extracted by prompting this model to answer ``What object?'' on the transcription $T$. Note that roBERTa can be replaced with other alternatives such as ChatGPT \cite{wu2023brief}, Gemini \cite{team2023gemini}, and Claude \cite{wu2023comparative}.

Once the desired object $T_O$ is extracted from the language instructions, the remaining task is to check whether the user specifies any object grasping preference composed of the holder $T_H$ and the holding part $T_P$. On qualitative assessment of user-provided utterances, we found that the part name is often the only other noun in the sentence, excluding the object and the references to the user such as ``I,'' ``me,'' and ``myself''. Thus, checking for the existence of the holding part $T_P$ can be done by searching for nouns that were not the object or the self-referents above.

Finally, when $T_P$ is set, we must also deduce $T_H$. We can acquire rich lexical clues for this task by constructing a directed graph of word dependencies using SpaCy \cite{spacy2}, and checking if pronouns referring to the user are ancestors of a verb. For example, in ``I want to grab,'' ``I'' is the subject ancestor of ``grab,'' implying that $T_H$ = human. When the pronoun is instead the descendant of the verb, (ie. ``give me'') then we invoke the reverse case, where $T_H$ = robot. When no such pronoun is found and the action verb is left with no ancestors, then the sentence is imperative and we can also infer $T_H$ = robot. Some sample queries and the corresponding parsed representations are shown in Table \ref{tab:sentences_objects_parts_holders}.

\begin{table*}[!ht]
\centering
\scalebox{0.9}{
\begin{tabular}{cccc}\toprule
\multirow{1}{*}{\textbf{Sentence} $(T)$}& \multirow{1}{*}{\textbf{Object} $(T_O)$} & \multirow{1}{*}{\textbf{Holding Part} $(T_P)$}
&\multirow{1}{*}{\textbf{Holder} $(T_H)$}\\
\midrule
\multirow{1}{*}{Give me the wooden hammer.}& \multirow{1}{*}{wooden hammer} & \multirow{1}{*}{N/A}
&\multirow{1}{*}{N/A}\\
\multirow{1}{*}{Hand over the cup to me.}& \multirow{1}{*}{cup} & \multirow{1}{*}{N/A}
&\multirow{1}{*}{N/A}\\
\multirow{1}{*}{Pass the toy plane over.}& \multirow{1}{*}{toy plane} & \multirow{1}{*}{N/A}
&\multirow{1}{*}{N/A}\\
\multirow{1}{*}{I want the orange}& \multirow{1}{*}{orange} & \multirow{1}{*}{N/A}
&\multirow{1}{*}{N/A}\\
\multirow{1}{*}{Hand me the mustard bottle by grabbing the tip.}& \multirow{1}{*}{mustard bottle} & \multirow{1}{*}{tip}
&\multirow{1}{*}{robot}\\
\multirow{1}{*}{Grab the screwdriver's shaft.}& \multirow{1}{*}{screwdriver} & \multirow{1}{*}{shaft}
&\multirow{1}{*}{robot}\\
\multirow{1}{*}{Deliver me the frying pan so I can hold the handle.}& \multirow{1}{*}{frying pan} & \multirow{1}{*}{handle}
&\multirow{1}{*}{human}\\
\multirow{1}{*}{I want to hold the apple by the stem.}& \multirow{1}{*}{apple} & \multirow{1}{*}{stem}
&\multirow{1}{*}{human}\\
\multirow{1}{*}{Give me the knife by its handle.}& \multirow{1}{*}{knife} & \multirow{1}{*}{handle}
&\multirow{1}{*}{human}\\
\bottomrule
\end{tabular}}
\caption{Our language parsing module extracts the object $T_O$, the holding part $T_P$, and the holder $T_H$ from the given the language transcription $T$. This table shows the results from some sample user sentences.} 
\label{tab:sentences_objects_parts_holders}
\end{table*}

\subsection{Multimodal Object Selection}\label{sec3sub4}
Given the result from the gaze detection module and language parsing model, our multimodel object selection module determines the user's object of interest and the corresponding holding part if specified.

For the language modality of our pipeline, we must locate the object in the scene given its name in a textual representation. Since we intend for our pipeline to generalize to an arbitrary set of objects, relying on a classification model that is tuned for a set of specific objects is not the optimal option. Instead, we leverage the Grounded Language-Image Pretraining (GLIP) \cite{li2022grounded} to serve as the interpreter of the natural language input. GLIP is conditioned on a large corpus of text and bounding box data, so it can support varied object names and descriptions from the users. For example, both ``spray bottle" and ``windex" are reasonable names that the users may use to refer to the common household cleaning tool, and these variants will be interpreted by GLIP as the same object. We use GLIP instead of other large referring segmentation approaches like \cite{LISA} \cite{lin2023wildrefer} \cite{margffoytuay2018dynamic} integrating full-sized language models due to its high result quality and low computiation cost.

\begin{algorithm}[hbt!]
\caption{Object\_Selection}\label{alg:object_selection}
 \hspace*{\algorithmicindent} \textbf{Input:$I_c$, $H_g$, $T_O, T_P, T_H$ } \Comment{scene image, gaze heatmap, desired object, hold part, holder}\\
 \hspace*{\algorithmicindent} \textbf{Output: $B_{O_u}$, $B_{O_p}$} \Comment{bounding box of the object of interest and the robot holding part} 
\begin{algorithmic}[1]
%\Require{$I_c$, $H_g$, $T_O, T_P, T_H$ \Comment{scene image, gaze heatmap, desired object, hold part, holder}}
%\Result {$B_{O_u}$, $B_{O_p}$ \Comment{bounding box of the object of interest and the robot holding part}}
\State $\{B_{O_i}\}_k \leftarrow \text{GLIP}(I_c, T_O)$ \Comment{all potential bounding boxes}\;
\State $\{S_{O_i}\}_k \leftarrow \text{IoU}(H_g, \{B_{O_i}\}_k)$ \Comment{calculate IoU scores}\;
\State $B_{O_u} \leftarrow \text{argmax}_{B_{O_i}}(\{S_{O_i}\}_k)$ \Comment{extract object of interest}\;
\If{$T_P != \emptyset$}
  \State $\{B_p\}_q \leftarrow \text{GLIP}(I_c, T_p)$ \Comment{find all potential parts}\;
    \State $B_{O_p} \leftarrow \text{argmax}_{B_p}(B_{O_u} \cap \{B_p\}_q)$ \Comment{locate the part}\;
    \If{$T_H = $ Robot}
        \State $B_{O_p} \leftarrow (B_{O_u} \cap B_{O_p})$ \Comment{robot grasp specified}\;
    \ElsIf{$T_H = $ Human}
        \State $B_{O_p} \leftarrow B_{O_u} - (B_{O_u} \cap B_{O_p})$ \Comment{user grasp specified}\;
    \EndIf
\EndIf
\State \Return $B_{O_u}, B_{O_p}$
\end{algorithmic}
\end{algorithm}

Algorithm \ref{alg:object_selection} reveals how we use the results from previous modules to extract the bounding box of the object of interest $B_{O_u}$ and that of the corresponding robot holding part $B_{O_p}$. The inputs of the algorithm contain the RGB image $I_c$, the gaze heatmap $H_g$, and the language parsing result tuple $(T_O, T_P, T_H)$. First, the textual form of the desired object $T_O$ is sent to the GLIP model along with the RGB image $I_c$ to generate the bounding boxes for all the potential objects, denoted as $\{B_{O_i}\}_k$ (Line 1). Then, for each bounding box $B_{O_i}$, an intersection over union (IoU) score $S_{O_i}$ is calculated by summing over the associated probability density values of all inside pixels from the gaze heatmap $H_g$, forming the score set $\{S_{O_i}\}_k$ (Line 2). The bounding box of the object of interest $B_{O_u}$ is selected to be the one with the largest IoU value (Line 3). In the cases where the user also specifies the holding part $T_P$ and holder $T_H$, a similar approach is used. This time, the GLIP model takes the holding part $T_P$ as the input and outputs the bounding box of all the potential parts $\{B_p\}_q$ (Lines 4-5). The algorithm then finds the corresponding part $B_{O_P}$ on the previously detected object of interest $B_{O_u}$ (Line 6). If the holder $T_H$ is the robot, the intersection between the object $B_{O_u}$ and the part $B_{O_P}$ is returned since that's the only region that can be grasped by the robot (Lines 7-8). However, when the holder is set to bet the human, our algorithm removes the detected part from the object bounding box to prevent the robot from accessing there (Lines 9-10). Finally, the bounding box of the object of interest $B_{O_u}$ and that of the robot holding part $B_{O_p}$ are returned (Line 11). In summary, our algorithm extracts the desired object and the robot holding part by jointly considering both the user's gaze and the language instructions.

\subsection{Robot Grasp Generation}\label{sec3sub5}
Once the bounding box of the desired object is found, the robot grasp generation method extracts its point cloud and outputs the best robot grasping pose for handover as shown in Algorithm \ref{alg:grasp_selection}. Our approach first finds the object's point cloud by projecting the region of $I_d$ that lies within $B_{O_u}$ using the camera intrinsic/extrinsic parameters (Line 1). Since depth measurements can be inaccurate along the edge of objects, we also remove outlier points that deviate too much from their neighbors. As the back surface of the object is not visible, we further use PoinTr \cite{poin_tr} model to infer the complete point cloud of the object (Lines 2-4). Then, the robot grasp generation is performed in two stages. In the first stage, to enforce a safe human-robot handover process, our method generates potential robot grasps that leave enough safety margin for the human hand when presenting the object of interest. While in the second stage, the best robot grasp is selected among all the candidates considering criteria like the grasp stability, the approach angle, and the minimal distance towards the human hand. Based on different grasping preference specifications, the pipeline varies slightly as follows:
\begin{enumerate}
    \item If the robot grasping preference is specified, Contact-GraspNet \cite{9561877} is utilized to generate a set of robot grasps on the point cloud of the specified object part $P_{pc}$. Contact-GraspNet is designed to predict diverse robust robot grasps for various objects by considering grasp geometry. Since the robot grasping part is instructed by the user, the resulting robot grasps should already avoid the area that the user intends to grasp, which results in a safe handover experience. Furthermore,  Contact-GraspNet evaluates the generated grasp configurations using a confidence score indicating the grasp stability. Thus, our method returns the robot grasp with the largest confidence score (Lines 5-6).
    \item If the human grasping preference is specified, a similar approach to the above case is used. Recall in Algorithm \ref{alg:object_selection}, when the holder is the human, $B_{O_p}$ is set to be the remaining part of the object after taking out the portion that will be accessed by the human. Thus, all potential robot grasps and their corresponding scores can be found by directly feeding $P_{pc}$ into Contact-GraspNet (Lines 5-6). 
    \item When neither human nor robot preference is specified, our algorithm first predicts socially-complaint human grasping positions respecting stability, safety, and social conventions for the robot grasp to avoid. Note that this human hand prediction approach is inspired by CoGrasp \cite{10160623}. However, instead of training the VAE-based human hand predictor from scratch, we directly employ GraspTTA \cite{grasp_tta} as it focuses on generating grasps by modeling the consistency between the object contact regions and hand contact points. GraspTTA produces realistic and functional grasps by ensuring the predicted hand contact points are either close to or touch the objects' contact regions. After engineering multiple possible hand positions, we take advantage of Contact-GraspNet to generate all feasible grasping poses (Line 9). At last, CoGrasp's scoring function is invoked to combine the outputs of GraspTTA and Contact-GraspNet - human hand prediction and robot grasps, respectively - to reevaluate the grasps based on their human suitability (Line 10). This algorithm combines two individual scores based on the distance and the approaching angle towards the human hand as explained below:
    
    \textit{Distance Measure ($S_d$)}: This metric estimates the distance between the robot and the predicted human hand. Using the point cloud of the robot gripper $PC_g$ and that of the human hand $PC_h$, the distance measure $S_d$ can be computed as the average sum of the pairwise Euclidean distance between them using Equation \ref{eq5}: 
    \begin{equation}\label{eq5}
        S_d(PC^g, PC^h) = \frac{\sum_{\substack{x\in PC^g \\ y\in PC^h}}\lVert x - y \rVert ^2}{|PC^g||PC^h|}
    \end{equation}

    \textit{Angle Measure ($S_a$)}: This value calculates the difference between the approaching angle of the robot gripper and that of the predicted human hand. Assume the approach vectors for the robot and the human hand are denoted as $a_g$ and $a_h$, respectively. Then the angle measure is formally defined as the inner product regarding $a_g$ and $a_h$ as follows:
    \begin{equation}
        S_a(a_g, a_h) = - (a_g \cdot a_h)
    \end{equation}

    We sum the distance measure $S_d$ and angle measure $S_a$ to serve as the final criteria $C$ for robot grasp generation. Among all the potential robot grasps, our module favors those with larger $C$ values indicating they approach the object from a different direction compared to the human hand while keeping a considerable distance to the hand at the grasp pose.

\end{enumerate}

Using the respective evaluation system for each case, the final robot grasp $G_{O_u} \in SE(3)$ is chosen to be the one with the highest score (Line 11). The resulting human-aware robot grasp is the most suitable one for the object of interest $O_u$ as it strictly obeys the user's preference if it is specified. Otherwise, the grasp accounts for it implicitly by generating the potential human grasps.

\begin{algorithm}[hbt!]
\caption{Grasp\_Generation}\label{alg:grasp_selection}
 \hspace*{\algorithmicindent} \textbf{Input:$I_d, B_{O_u}, B_{O_p}, T_H$} \Comment{color image, depth image, object bounding box, part bounding box, holder}\\
 \hspace*{\algorithmicindent} \textbf{Output: $G$} \Comment{best grasp}
 \begin{algorithmic}[1]
%\KwData{$I_d, B_{O_u}, B_{O_p}, T_H$ \Comment{color image, depth image, object bounding box, part bounding box, holder}}
%\KwResult{$G$ \Comment{best grasp}}
%$f_s \gets \text{CoGraspScoring()}$\;
\State $pc \leftarrow $PC\_Extractor$(B_{O_u}, I_d)$\;
\State $pc' \leftarrow$ PoinTr($pc$) \Comment{point cloud completion}\;
\State $O_{pc} \leftarrow$ PC\_Extractor($B_{O_u}, pc'$) \Comment{completed object}\;
\State $P_{pc} \leftarrow$ PC\_Extractor($B_{O_p}, pc'$) \Comment{completed part}\;
\If{$T_H = $ Robot or $T_H = $ Human}
    \State $\{grasp, score\}_m \leftarrow$ Contact\_GraspNet($P_{pc}$)
\Else
    \State $\{hand\}_k \leftarrow$ GraspTTA($O_{pc}$)\;
    \State $\{grasp\}_q \leftarrow$ Contact\_GraspNet($O_{pc}$)\;
    \State $\{grasp, score\}_m \leftarrow \text{CoGrasp}(\{grasp\}_q, \{hand\}_k)$
\EndIf
\State $G_{O_u} \gets argmax_{grasp}{\{score\}}_m$\;
\State \Return $G_{O_u}$
\end{algorithmic}
\end{algorithm}

\subsection{Handover Motion Generation}\label{sec3sub6}
This section introduces our handover motion generation module that generates smooth robot movements for object handover leveraging the Riemannian Motion Policy (RMP) \cite{Ratliff2018RiemannianMP}. RMP allows us to combine locally reactive policies for obstacle avoidance and trajectory smoothness to ensure fluid and predictable movements, which are crucial for safe human-robot interaction. It also minimizes instability as objects are being retrieved and handed to the user. 

As suggested by \cite{Ratliff2018RiemannianMP}, we first incorporate a local policy that is defined by the following second-order differential equation:
\begin{equation}
    \label{eq:f}
    f(x, \dot{x}) = \ddot{x} = \kappa \cdot s(x,x_g) - \Omega \dot{x}
\end{equation}
Here, $\kappa$ and $\Omega$ are scaling hyperparameters, with $x$ and $x_g$ representing the current position and the goal position, respectively. When the robot is grasping the object, the target pose is $G_{O_{u}}$. On the other hand, at the object handover phase, the target is set to be a fixed pose near the user $G_{U} \in SE(3)$ for easier access. The number of dots on $x$ represents the order of the derivative with respect to time; for example, $\dot{x}$ denotes the velocity. The policy mentioned in the above equation outputs an acceleration proportional to the soft normalization distance $s (\cdot)$ between the current and target pose and inversely proportional to the velocity, preventing extreme action. We construct an RMP from this, which is defined as a set with the acceleration function and the Riemannian metric. %The Riemannian metric scales the accelerations and velocities based on the direction and the space.

When the acceleration is calculated using Equation \ref{eq:f} in the task space $W$, the next step is to map it to the robot's 6-DOF joint space $Q$ using a pull function. The pull function utilizes the Jacobian matrix $J$ associated with the robot's forward kinematics function $\phi$ to produce the joint space acceleration based on the output from the task space as shown in Equation \ref{eq:pull_push}:
\begin{equation}
    \label{eq:pull_push}
    \begin{split}
        & \text{pull}_{\phi}(^{W}(f, A)) =~^Q(J^+f, J^\top AJ) \\
        % & \text{push}_{\phi}(^Q(h, A)) = ^X(Jh, (J^+)^TAJ^+)
    \end{split}
\end{equation}

The pull operation transfers an RMP defined in the 3D Euclidean space to that of the robot’s joint space. The  $A$ is a Riemannian metric defined at every point in the manifold. Since our task is performed in the 3D Euclidean space, setting the Riemannian metric $A$ to an identity matrix makes the defined length to be the Euclidean norm. Furthermore, $J^\top$ and $J^+$ are the transpose and the pseudoinverse of the Jacobian, respectively.

Finally, at each time step during the object grasping and handover phases, our motion generation method leverages the above-mentioned pull function to generate the corresponding joint space acceleration and velocity, which eventually leads to a new robot configuration. By iterating through the time steps, a sequence of robot arm configurations $\{q\}_m$ is constructed to perform smooth motions for object grasp and handover.

\subsection{Workflow}\label{sec3sub7}
With all the modules defined and discussed, the complete pseudocode of our proposed multimodal pipeline for precise object handover is shown in Algorithm \ref{alg:whole_pipeline}. At first, the robot takes an RGB-D image $I_c, I_d$ of the whole scene and sends it to the user's monitor (Line 1). The user then utilizes eye gaze and natural language prompts to specify an object of interest $O_u$ for handover while our system records the corresponding facial image stream $\{I_f\}_n$ and language transcription $T$ for further processing (Lines 2-3). In order to project the high-level user instructions to an object in the scene, the gaze module is designed to first figure out where the user is staring at on the image $I_c$ and then generate a heatmap $H_g$ on top of it indicating the probability of each pixel being the user's eye focal point (Line 4). At the same time, our language parser examines the user instruction transcription $T$, resulting in the textual form of the user's object of interest $T_O$, the holder $T_H$, and the holding part $T_P$ (Line 5). With the pre-processing of the user's visual and verbal evidence finished, the multimodal object selections method consumes the outputs from the upstream modules and outputs bounding boxes of the user's desired object and the holding part $B_{O_u} \text{ and } B_{O_p}$ (Line 6). In the next step, the best human-aware robot grasp $G_{ou}$ accounting for the user's preference, the human suitability, and the geometric stability is returned by our grasp generation approach (Line 7). Finally, our pipeline leverages the handover motion generation method to move the robot to the desired grasping pose, obtain the object, and hand it over safely to the user at a fixed pose $G_U \in SE(3)$. (Lines 8-9).

\begin{algorithm}[hbt!]
\caption{Multimodal Object Handover}\label{alg:whole_pipeline}
\begin{algorithmic}[1]
\State $I_c, I_d \leftarrow $ Robot\_Hand\_Camera() \Comment{send image to user}\;
\State $\{I_f\}_n \leftarrow$ Facial\_Camera($I_c$, user)\Comment{record user images}\; 
\State $T \leftarrow$ Microphone(user) \Comment{record user's instruction}\;
\State $H_g \leftarrow $Gaze\_Detection($\{I_f\}_n$) \Comment{user's gaze heatmap}\;
\State $(T_O, T_P, T_H) \leftarrow$ Language\_Parsing(T)\;
\State $B_{O_u}, B_{O_p} \leftarrow$ Object\_Selection($I_c, H_g, T_O, T_P, T_H$)\;
\State $G_{O_u} \leftarrow$ Grasp\_Generation($I_d, B_{O_u}, B_{O_p}, T_H$)\;
\State $\{q\}_m \leftarrow$ Motion\_Generation($G_{O_u}$) \Comment{Object grasp}\;
\State $\{q\}_n \leftarrow$ Motion\_Generation($G_U$) \Comment{Object handover}\;
\end{algorithmic}
\end{algorithm}

\section{Experiments}\label{Sec4}
We conduct a series of real-world experiments involving randomly selected users with no prior bias to evaluate the performance of the essential subsystems and the overall framework. The tested subsystems include gaze detection, object extraction, and grasp generation. Each test comes with a set of uniquely designed metrics to assess the effectiveness of the module.

\subsection{Environment Setup}\label{sec4sub1}
\label{subsec:envstp}
\begin{figure*}[ht]
    \centering
    \centering
    \includegraphics[width=.49\textwidth]{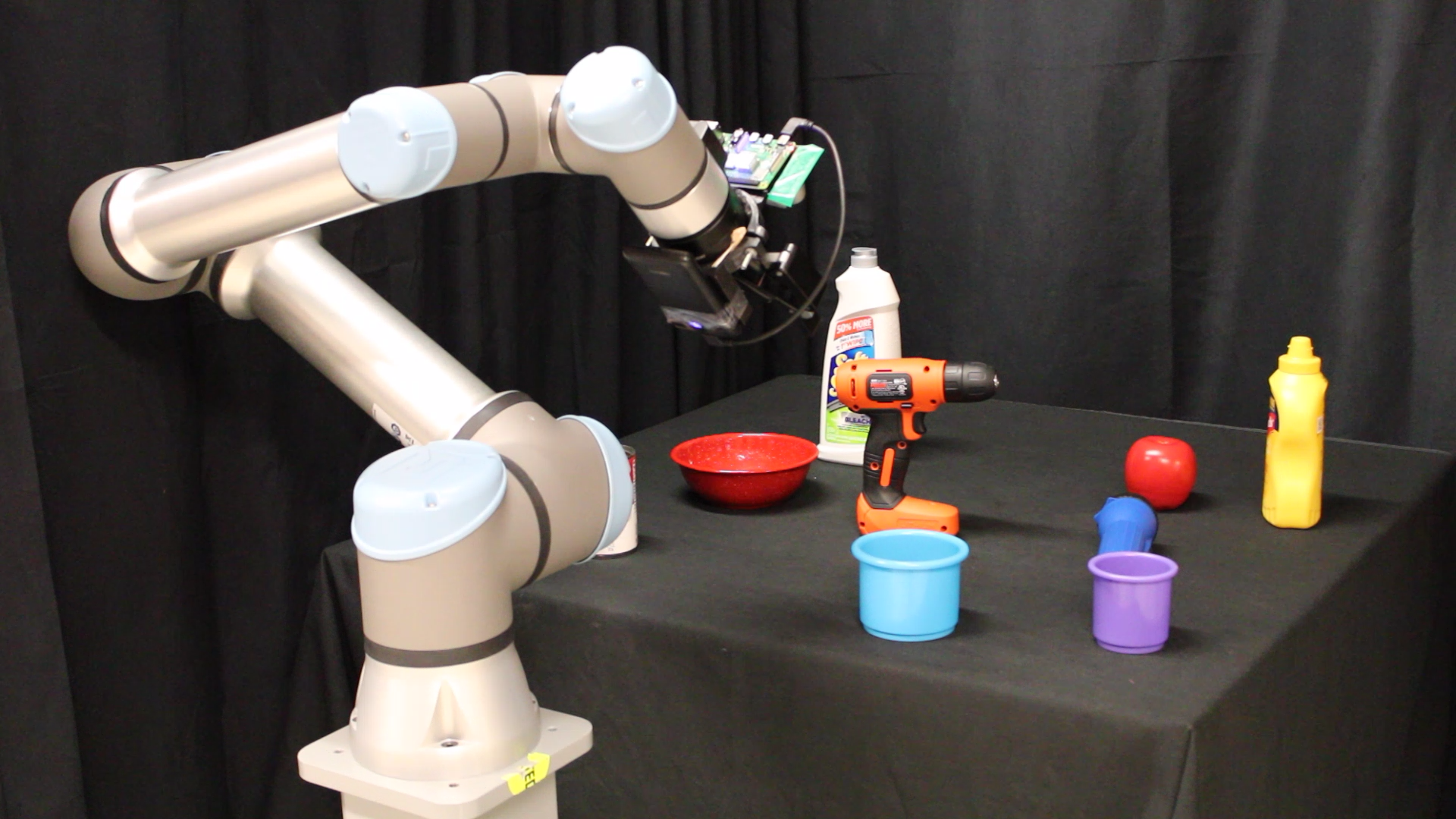}\hfill
    \includegraphics[width=.49\textwidth]{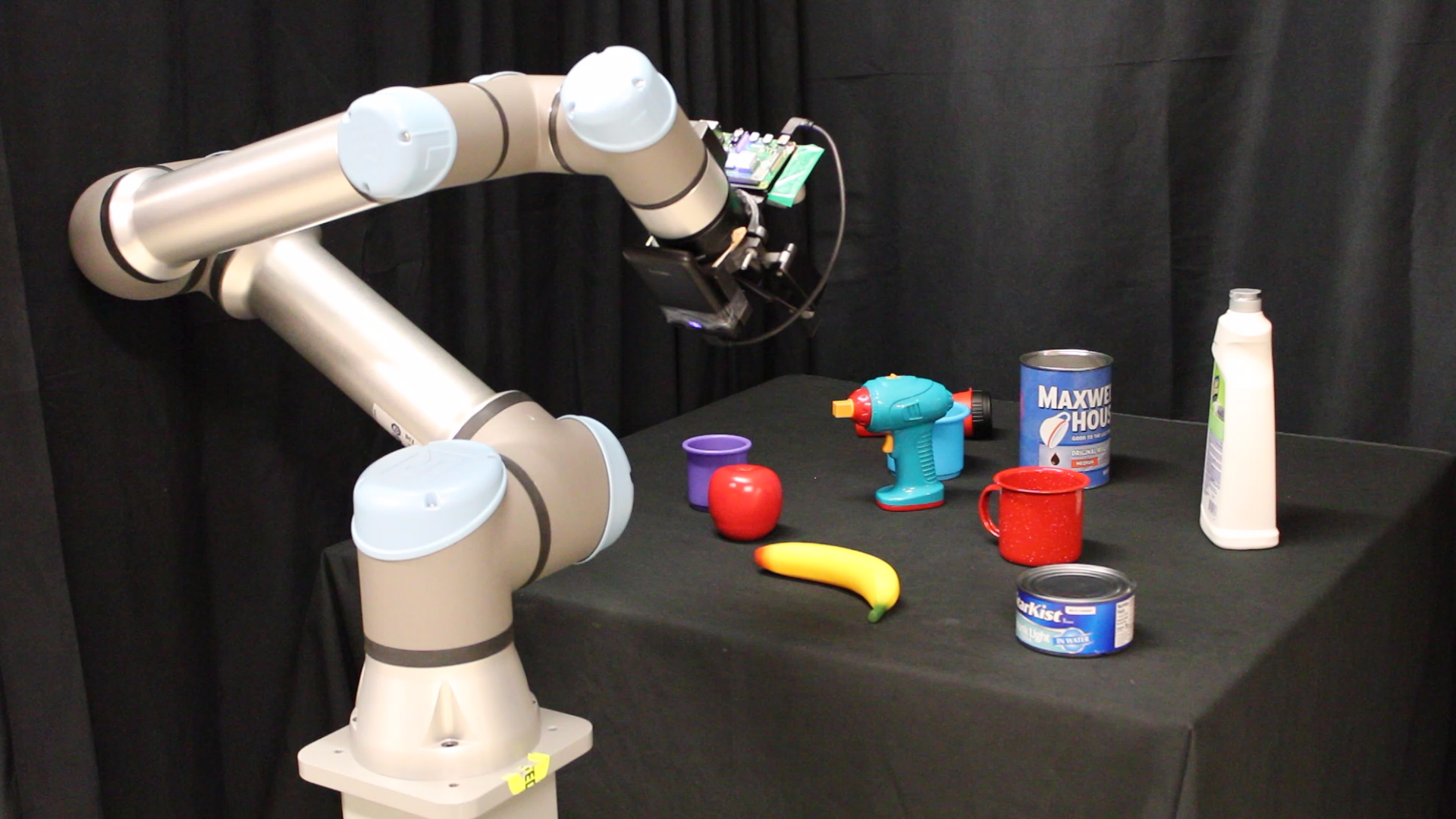}
    \caption{This figure shows two sample tabletop setups in our real experiments. For each environment, 8-10 arbitrarily selected household objects from the YCB dataset are placed randomly on the table. A UR5e robot arm with a Robotiq 2F-85 gripper and an Intel RealSense D435i camera attached is placed in front of the table as the agent for scene capture and object handover.}
    \label{fig:tabletop_figure}
\end{figure*}

As illustrated in Fig.~\ref{fig:tabletop_figure}, in our experimental environment, we set up a table of dimensions (50 cm, 50 cm) in width and length with a set of 8-10 arbitrarily selected YCB \cite{YCBDataset} household items placed randomly on top. A UR5e robot arm manipulator (Universal Robotics, Denmark) with a 2F-85 gripper attached (Robotiq, Canada) is deployed in front of the table to serve as the handover agent. On the robot arm, a RealSense D435i (Intel, USA) camera is strategically mounted onto the wrist link to capture the RGB-D image of the environment. The resulting image is relayed back to the user's monitor through a Raspberry Pi (Model 3 B+, Raspberry Pi Foundation, UK), which is also mounted onto the robot arm's wrist. On the user end, to capture the user's facial image stream, a RealSense D435i is mounted on top of the monitor display with a clear view of the user's face and shoulders. The duration of the Gaze capture is variable based on the length of the natural language input. We use a Conda environment to manage and containerize all required packages. The key libraries include GraspTTA, ContactGraspNet, Gaze, and GLIP. The codebase is entirely written in Python. The user's laptop captures only verbal instructions, gaze data, and images of the scene. These inputs are transmitted to a server via the SSH protocol, where computationally intensive tasks such as human hand pose estimation, vision-language referral, and grasp generation are performed. Once processed, only the final robot grasp is sent back to the laptop, which then interfaces with the UR5e robot via UR’s RTDE client library.

\subsection{Subsystem Evaluation}
\label{subsec:eval}
Our subsystem evaluations are organized into 4 sections, each focused on a specific part of the pipeline. To ensure thorough assessment over a diverse range of scenarios, we built 50 randomized scenes using the YCB \cite{YCBDataset} object dataset. This dataset was chosen because it has been widely used in previous research and in real-world scenarios. Utilizing this dataset allowed us to build complex real-world-like scenes featuring multiple instances of the same object with slight variations in size or color to further reveal the strong performance of our pipeline.
\subsubsection{Gaze Detection} 
\label{subsec:gaze_exp}
This test examines whether our gaze detection method results in selecting the correct user's object of interest in the image. In the trials, we generate 50 scenes of 9 objects each, where the location and type of object is chosen at random. In addition, we randomly select 2 objects in each set of 9 to be the target objects. Then, each of three participants are presented with each scene one at a time and informed of the target object $O_u$ they need to focus on. To ensure gaze is properly measured, a blind is used to prevent users from directly observing the scene, requiring them to view the scene as an RGB image $I_c$ through the monitor. Once the user's facial image stream $\{I_f\}_n$ is collected and sent to our gaze module, we record the resulting pixel of interest $(\lambda, \mu)$ and the heatmap $H_g$. The following three metrics are used to evaluate the gaze result with the help of the ground truth object bounding box $B_{O_u}$:
\begin{itemize}
    \item \textbf{Success Rate (SR):} For each object, a sum of the probability density is calculated. Then, the object with the highest probability is predicted as the user's chosen object of interest $O_u'$.
    \item \textbf{Intersection over Union (IoU):} First, the heatmap $H_g$ is discretized by removing regions with a probability density value $< 0.01$. Then, the IoU is calculated by summing the probability density values of $H_g$ within $B_{O_u'}$, followed by dividing by the area of the union $B_{O_u'} \cup H_g$.
    \item \textbf{Mean Squared Error (MSE):} When the user's pixel of interest is not within the correct object's bounding box, the MSE in pixel units is computed to measure shifting error between the gaze point (at the center of $H_g$) and $B_{O_u}$.
\end{itemize}

\begin{table}[htbp]
    \centering
    \scalebox{0.7}{
    \begin{tabular}{lcccc}
        \toprule
        & \textbf{Object} & \textbf{ SR $(\%)$ $\uparrow$} &\textbf{MSE (pixel) $\downarrow$} & \textbf{IoU $(\%)$ $\uparrow$} \\
        \midrule
        \multirow{2}{*}{\textbf{Participant 1}} & Object 1 & 76\% & 0.55 & 36.4\% \\
        & Object 2 & 78\% & 1.64 & 35.9\%\\
        \midrule
        \multirow{2}{*}{\textbf{Participant 2}} & Object 1 & 94\% & 0.33 & 39.6\% \\
        & Object 2 & 88\% & 1.28 & 37.3\%\\
        \midrule
        \multirow{2}{*}{\textbf{Participant 3}} & Object 1 & 78\% & 1.28 & 34.9\%\\
        & Object 2 & 82\% & 1.47 & 35.6\%\\
        \midrule
        \textbf{Overall} &  & 83\% & 1.09 & 36.6\%\\
        %\textbf{Standard Deviation} & - & 0.36 & 3.91 & 16.2\%\\
        \bottomrule
    \end{tabular}}
    \caption{This table reveals our gaze detection results. Across all participants in this evaluation, we collected data on a total of 300 gazes. Overall, 249/300 such gazes successfully identified the ground truth target object. In our full pipeline, we further augment the object selection accuracy by introducing language based user inputs.}
    \label{tab:gaze_results}
    %\vspace{-0.1in}
\end{table}
\begin{figure*}[ht]
    \centering
    \centering
    \includegraphics[trim = {0.7cm 0cm 1cm 0cm}, clip, width=1.0\textwidth]{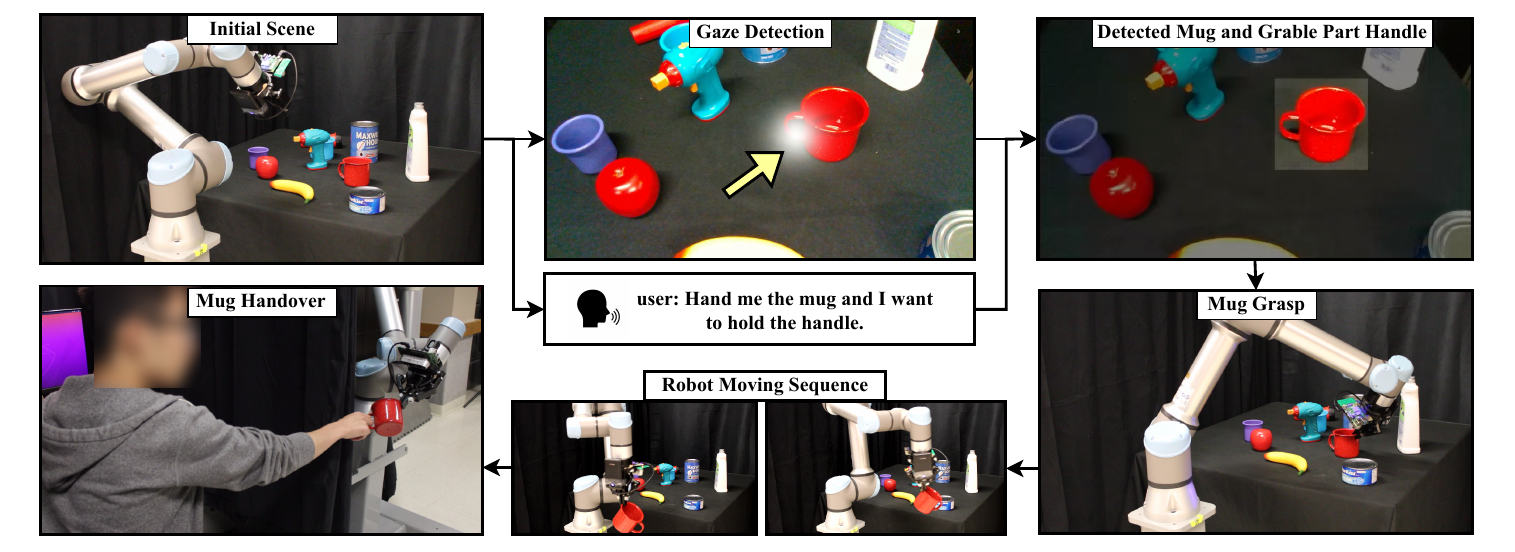}
    \caption{This figure demonstrates how our pipeline works in real-world experiments. At first,  the in-hand camera captures the cluttered environment and sends it to the user's monitor. The second image on the top row is the image displayed on the user's monitor with the bright dot indicated by the yellow arrow being the real-time detected gaze. The user's eyesight focuses on the red mug in this test. At the same time, the microphone records the user's verbal instructions as ``Hand me the mug and I want to hold to handle''. By jointly considering all user inputs, our system figures out the correct object of interest and the corresponding grasping preference. After generating the grasping pose that strictly obeys the user's instruction by holding the mug on the rim, the robot then moves to obtain the mug and finally hands it over safely to the user, leaving the handle for the user to grasp. }
    \label{fig:real_exp_figure_1}
\end{figure*}

The result is shown in Table \ref{tab:gaze_results}. The gaze module successfully captures the correct object a large majority of the time---at least 76\% and as high as 96\%. We find that users tend to focus on the specific subregions of an object that they intend to hold rather than the direct center of the object, which could contribute to the lower IoU scores that range from $35\% - 40\%$. Finally, the low MSE values demonstrate that, although gaze detection has occasional failures, these errors are on average not more than 1.64 pixels from the edge of the bounding box. We observed that such failures are more due to cluttered objects than a fundamental inaccuracy in the gaze detection itself. This allows for additional language-based input from the user to boost object selection accuracy, as shown later in Figure \ref{tab:obj_selection}.

\subsubsection{Gaze Error Margin Evaluation}
In this section, we investigate the minimum gap between identical objects required by our gaze system to work accurately. To evaluate this, we conducted an additional set of experiments. Involving three participants, we arranged two identical objects and measured the minimum distance between objects required for our gaze system to function accurately. We found that the minimum gap threshold depends on the size of the objects. 

We categorized the objects into three sizes based on their widths: small (less than 1 cm), medium (between 2 cm and 4 cm), and large (greater than 8 cm). During the trials, participants were instructed to use gaze tracking to select one object from the identical pair. Initially, we placed the objects 10 cm apart (measured from surface to surface) and then incrementally reduced the distance until a participant could no longer successfully select the target object, with success defined as described in Section \ref{subsec:gaze_exp}.

Our results show that the minimum gap needed for precise selection of objects with our gaze system is 5 cm for large objects (e.g., flashlight), 4.5 cm for medium objects (e.g., clamps), and 3.75 cm for small objects (e.g., pencil eraser). The relatively short distances demonstrate the effectiveness of our gaze system in distinguishing identical objects.

\subsubsection{Object Selection} 
The test is designed to check how well our object selection approach performs in different settings. Since the object selection module is built based on both gaze detection and language parsing, we set up a comprehensive test of it considering the following three scenarios:
\begin{itemize}
    \item {
        The object selection is only based on gaze detection. This experiment is similar to the stand along test for gaze detection. For each of the 50 testing environments, 3 users are asked to focus on their eyesight on a specific object. Then, the previously described maximum probability method from the heatmap is used to select the correct object.
    }
    \item {
        The object selection is only based on parsed user language instructions. For each of the 50 testing environments, we randomly select a target object. Then, using only its name, GLIP produces bounding box candidates for the target object. The box with the highest GLIP confidence is chosen as the final prediction.
    }
    \item {
        The last scenario is our full object selection module combining both gaze detection and language parsing. First, GLIP produces multiple candidate bounding boxes based on the language prompt. Then, the bounding box that has the highest IoU with the heatmap is selected as the chosen object.
    }
\end{itemize}

\begin{figure*}[ht]
    \centering
    \centering
    \includegraphics[trim = {0.7cm 0cm 1cm 0cm}, clip, width=1.0\textwidth]{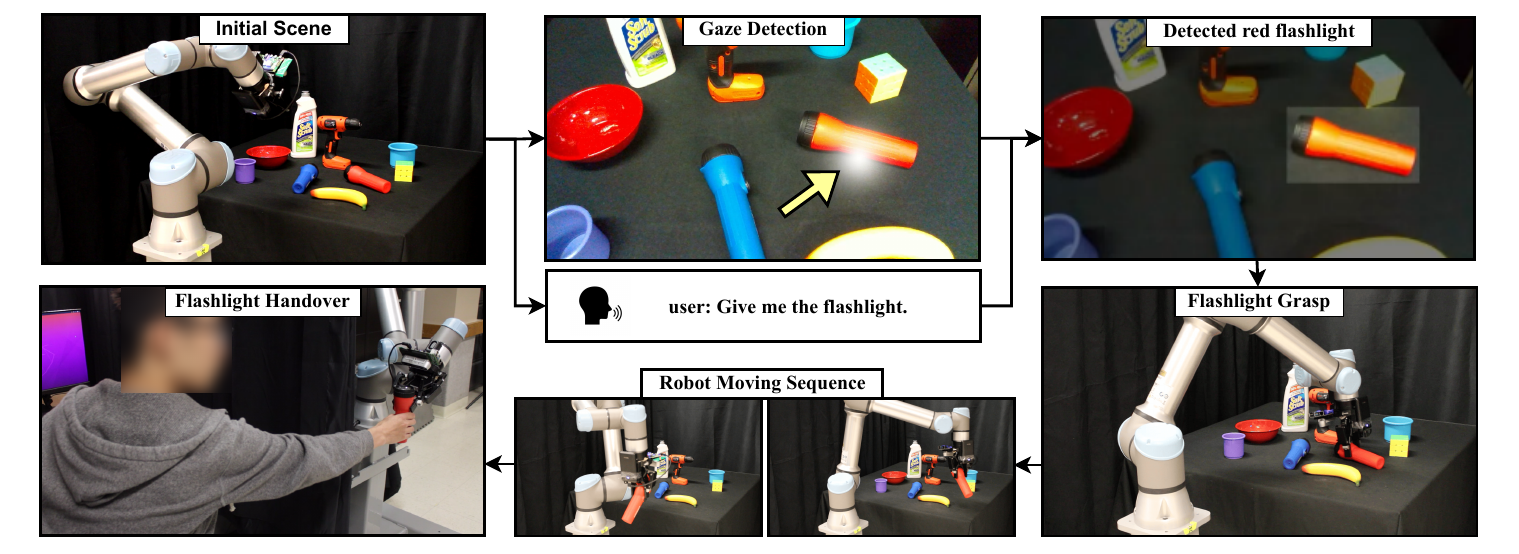}
    \vspace{-10pt}
    \caption{The images showcase the workflow of another real experiment. The environment is captured by the robot's in-hand camera and sent to the user's monitor. There are two flashlights in the scene and the user seeks to obtain the red one, indicated by the detected gaze in the second image on the first row. However, the language instruction given by the user is vague simply saying ``give me the flashlight''. For the sake of our multimodal pipeline, the bounding box of the red flashlight is correctly extracted, shown in the top right image. Since the user does not give out any holding preference, our human-aware grasp generation module proposes a grasping pose that is near the front end of the flashlight, leaving a large space at the middle and rear end for the user. As a result, the user successfully and safely retrieves the desired red flashlight.}
    \label{fig:real_exp_figure_2}
\end{figure*}

% For each of the individual tests, we record the detected object of interest $O_u'$ and compare it with the ground truth $O_u$. Table \ref{tab:obj_selection} reveals the average success rate of all test settings. As expected, we observe that the user's language instructions are not enough for the module to select the right object. The reason behind this lies in the fact that our complex testing environment is cluttered and often involves multiple copies of the same object, i.e., a red flashlight and a blue one. In these cases, if the user only specifies the object name flashlight without any additional detail, the model's success rate is at most 50$\%$. On the other hand, the test with gaze only shares a similar success rate with the stand-alone test described in the previous part. Finally, the success rate rises to 96$\%$ for the complete object selection method. Although the language instruction itself is not enough, it helps to improve the performance significantly when combined with gaze detection. For example, in our experiments, when the user fails to look directly at the desired object, gaze detection will struggle to identify the target object. However, with the help of the parsed language instructions, the non-target objects are filtered out, leaving only the true object of interest. 

For each of the individual tests, we record the detected object of interest $O_u'$ and compare it with the ground truth $O_u$. Table \ref{tab:obj_selection} presents the average success rate under four different settings. As expected, relying solely on language instructions yields a low selection accuracy (48\%), especially because our cluttered testing environments often contain multiple copies of the same object (e.g., a red flashlight and a blue one). In contrast, using only gaze achieves an 82.67\% success rate, similar to the stand-alone gaze evaluation above.

Crucially, combining language with gaze boosts the success rate to 96.23\%. We verify that this improvement is statistically significant at $5\%$ significance level in Table \ref{tab:obj_selection_evidence}: performing an independent Z-test to compare gaze alone (82.67\%) versus gaze and language (96.23\%) yields a z statistic of -3.06, which is less than , while comparing language alone (48\%) versus gaze and language (96.23\%) yields a z statistic of -5.94. Thus, although language instructions alone do not suffice in cluttered scenes, they significantly enhance object selection when combined with gaze. For example, if the user initially doesn't look directly at the desired object, gaze detection alone may struggle; however, language constraints help filter out non-target objects, recovering the true object of interest.

\begin{table}[htbp]
    \centering
    \begin{tabular}{l|cc}
        \toprule
        \textbf{Object Selection}& \textbf{Without Gaze} & \textbf{With Gaze}\\
        \midrule
        \textbf{Without Language} & N/A & 82.67\%\\
        \textbf{With Language} & 48\% & 96.23\%\\
        \bottomrule
    \end{tabular}
    \caption{Object selection accuracy across four settings. Because multiple copies of the same item are present, language alone (48\%) struggles to identify the correct object. Gaze alone performs better (82.67\%), but combining gaze with language achieves the highest accuracy (96.23\%).}
    \label{tab:obj_selection}
\end{table}

\begin{table}[htbp]
    \centering
    %\vspace{13pt}
    \scalebox{0.74}{
    \begin{tabular}{l|cc}
        %\centering
        \toprule
        \textbf{Independent Z-Test} & \textbf{Z Statistic} & \textbf{Z Critical Value ($5\%$)}\\
        \midrule
        Gaze vs.\ Gaze + Language & -3.06 & -1.645\\
        Language vs.\ Gaze + Language & -5.94 & -1.645\\
        \bottomrule
    \end{tabular}}
    \caption{The independent Z-test confirms that the combined approach significantly improves selection accuracy compared to using either modality independently.}
    \label{tab:obj_selection_evidence}
    \vspace{-0.01in}
\end{table}

%As expected, the experiments revealed the combination of Gaze and natural language input to perform best. Interestingly though, the addition of natural language input to Gaze fine-tunes the object selection process. Furthermore, as suggested by scenarios with Gaze, it seems that the impact of verbal input seems to be subtle, but significant.

\subsubsection{Grasp Generation} This test aims to check whether a reasonable robot grasp pose is generated when the user gives out specific holding preferences. In the experiments, we use 50 randomly generated scenes, and in each scene, two objects were randomly selected for grasp generation. For each of the 2 selected objects in the 50 scenes, we evaluate the correctness of the predicted grasping poses. This grasp generation is described in Algorithm \ref{alg:grasp_selection}, and can follow one of three scenarios:
\begin{itemize}
    \item Only the robot's grasping preference is stated. A success is counted whenever the robot grasp falls within the region of the robot's preference part.
    \item Only the user's grasping preference is stated. A success is counted when the robot's grasp falls outside of the user's preferred grasping part.
    \item No preference is contained in the user's verbal instruction. We count a success if the robot grasp leaves room for the user to hold the object by its standard grasping area.
\end{itemize}

We define the standard grasping areas as conventional points of contact that must be left clear during the handover process for the user. We base our ground truth standard grasping areas on past studies that define them from real demonstrations of humans interacting with objects \cite{Zhu2021TowardHG} \cite{Lin2015RobotGP}. For example, scissors are often held by looping the thumb and index finger through the holes. In the case when no grasping preference is given, the robot grasps are generated to avoid the predicted hand pose from the GraspTTA model. If the selected robot grasp occupies the standard grasping area, it is considered a failure.

\begin{table}
\centering
\scalebox{0.8}{
\begin{tabular}{ccc}\toprule
%\multirow{2}{*}{\textbf{Text Specification}}&\multicolumn{2}{c}{\textbf{w/o Hand Prediction}} & \multicolumn{2}{c}{\textbf{w/ Hand Prediction}}\\ \cmidrule{2-5}
%&\multicolumn{1}{c}{w/o Human}&\multicolumn{1}{c}{w/ Human}&\multicolumn{1}{c}{w/o Human}&\multicolumn{1}{c}{w/ Human}\\
\textbf{Language Specification} & \textbf{Without Human} & \textbf{With Human}\\
\midrule
\multirow{1}{*}{\textbf{Without Robot}}& \multirow{1}{*}{$76\%$} & \multirow{1}{*}{$92\%$}\\
\multirow{1}{*}{\textbf{With Robot}} &\multirow{1}{*}{$92\%$} & \multirow{1}{*}{N/A}\\
 \bottomrule
\end{tabular}}
\caption{Analysis of the robot grasp generation accuracy. Detailed grasp specification helps to improve the overall performance while hand prediction allows it to be more robust.} 
\label{tab:grasp_select_results}
\vspace{-0.1in}
\end{table}
\iffalse
\begin{table}[htbp]
    \centering
    \begin{tabular}{|p{1.5cm}|cc|cc|}
        \hline
        \multirow{2}{*}{\parbox{1.5cm}{\centering\textbf{Text Specification}}} & \multicolumn{2}{c|}{\textbf{W/o Hand Prediction}} & \multicolumn{2}{c|}{\textbf{W/ Hand Prediction}} \\ \cline{2-5} 
                                    & \multicolumn{1}{c|}{W/o Human} & W/ Human & \multicolumn{1}{c|}{W/o Human} & W/Human \\ \hline
        \multicolumn{1}{|c|}{W/o Robot} & \multicolumn{1}{c|}{64\%}      & 92\%    & \multicolumn{1}{c|}{76\%}      & 92\%    \\
        \multicolumn{1}{|c|}{W/ Robot}  & \multicolumn{1}{c|}{76\%}      & N/A& \multicolumn{1}{c|}{92\%}      & N/A\\ \hline
    \end{tabular}
    \caption{Grasp Selection}
    \label{tab:grasp_select_results}
\end{table}
\fi

The test results showcasing the successful grasping rate are summarized in Table \ref{tab:grasp_select_results}. From the results, it is clear that detailed specification helps to improve the generated grasps. The language instruction provides a very informative hint to the grasp prediction module regarding the valid regions. For example, if the user requests a part to be left free, then the system knows the ground truth for the user's preferred grasp and there is no need to run human grasp prediction. Thus, the language instruction eliminates a great source of variance in the grasp generation output since the grasp trial declares success as long as the gripper obeys the part constraint. When the language is not specified, the grasp accuracy drops. This is primarily caused by the fact that the hand predictions generated by the GraspTTA model \cite{grasp_tta} are not well-aligned with the standard human hand grasping area. However, the performance can be enhanced when GraspTTA is replaced with better human hand estimator in the future. 

\subsubsection{Grasp Stability} This test focuses on the quality of the generated robot grasping poses on various YCB objects. We conduct the experiment by checking whether the grasp generated by our approach can lift the object and hold it until it stops oscillating and becomes completely still. To make the test more comprehensive, different parts are specified by the user for various objects if possible. For example, the user could prefer the hold the mug by the handle or by the rim. However, the part specified test is ignored for objects that no grasping pose can be stated as a preference, such as pears and strawberries. For each object with a specific grasp setting, 3 trials are conducted to remove the noise and randomness in the real-world scenario, leading to more promising and reliable results.
\begin{table}
\centering
\scalebox{0.66}{
\begin{tabular}{ccccccccc}\toprule
\multirow{3}{*}{\textbf{Objects}}&\multicolumn{4}{c}{\textbf{No Part Specified}} & \multicolumn{4}{c}{\textbf{Part Specified}}\\ \cmidrule{2-9}
&\multicolumn{4}{c}{Trials}&\multicolumn{4}{c}{Trials}\\
\cmidrule{2-9}
&\multirow{1}{*}{$\#1$}&\multirow{1}{*}{$\#2$}&\multirow{1}{*}{$\#3$}&\multirow{1}{*}{Success Rate}&\multirow{1}{*}{$\#1$}&\multirow{1}{*}{$\#2$}&\multirow{1}{*}{$\#3$}&\multirow{1}{*}{Success Rate}\\
\midrule

 Bowl & \multirow{1}{*}{\cmark}&\multirow{1}{*}{\cmark}&\multirow{1}{*}{\cmark}&\multirow{1}{*}{$3\slash 3$}&\multirow{1}{*}{\cmark}&\multirow{1}{*}{\cmark}&\multirow{1}{*}{\cmark}&\multirow{1}{*}{$3\slash 3$} \\

  Cup & \multirow{1}{*}{\cmark}&\multirow{1}{*}{\cmark}&\multirow{1}{*}{\cmark}&\multirow{1}{*}{$3\slash 3$}&\multirow{1}{*}{\cmark}&\multirow{1}{*}{\cmark}&\multirow{1}{*}{\cmark}&\multirow{1}{*}{$3\slash 3$} \\

   Flashlight & \multirow{1}{*}{\cmark}&\multirow{1}{*}{\cmark}&\multirow{1}{*}{\cmark}&\multirow{1}{*}{$3\slash 3$}&\multirow{1}{*}{\cmark}&\multirow{1}{*}{\cmark}&\multirow{1}{*}{\cmark}&\multirow{1}{*}{$3\slash 3$} \\

   Banana & \multirow{1}{*}{\cmark}&\multirow{1}{*}{\cmark}&\multirow{1}{*}{\cmark}&\multirow{1}{*}{$3\slash 3$}&\multirow{1}{*}{\cmark}&\multirow{1}{*}{\cmark}&\multirow{1}{*}{\cmark}&\multirow{1}{*}{$3\slash 3$} \\

   Drill & \multirow{1}{*}{\cmark}&\multirow{1}{*}{\cmark}&\multirow{1}{*}{\cmark}&\multirow{1}{*}{$3\slash 3$}&\multirow{1}{*}{\cmark}&\multirow{1}{*}{\cmark}&\multirow{1}{*}{\cmark}&\multirow{1}{*}{$3\slash 3$} \\

   Mug & \multirow{1}{*}{\cmark}&\multirow{1}{*}{\cmark}&\multirow{1}{*}{\cmark}&\multirow{1}{*}{$3\slash 3$}&\multirow{1}{*}{\cmark}&\multirow{1}{*}{\xmark}&\multirow{1}{*}{\cmark}&\multirow{1}{*}{$2\slash 3$} \\

   Chief Can & \multirow{1}{*}{\cmark}&\multirow{1}{*}{\cmark}&\multirow{1}{*}{\cmark}&\multirow{1}{*}{$3\slash 3$}&\multirow{1}{*}{\cmark}&\multirow{1}{*}{\cmark}&\multirow{1}{*}{\cmark}&\multirow{1}{*}{$3\slash 3$} \\

   Screw Driver & \multirow{1}{*}{\cmark}&\multirow{1}{*}{\cmark}&\multirow{1}{*}{\cmark}&\multirow{1}{*}{$3\slash 3$}&\multirow{1}{*}{\cmark}&\multirow{1}{*}{\cmark}&\multirow{1}{*}{\xmark}&\multirow{1}{*}{$2\slash 3$} \\

   Scissors & \multirow{1}{*}{\xmark}&\multirow{1}{*}{\cmark}&\multirow{1}{*}{\cmark}&\multirow{1}{*}{$2\slash 3$}&\multirow{1}{*}{\cmark}&\multirow{1}{*}{\cmark}&\multirow{1}{*}{\xmark}&\multirow{1}{*}{$2\slash 3$} \\

   Fish Can & \multirow{1}{*}{\cmark}&\multirow{1}{*}{\cmark}&\multirow{1}{*}{\cmark}&\multirow{1}{*}{$3\slash 3$}&\multirow{1}{*}{\cmark}&\multirow{1}{*}{\cmark}&\multirow{1}{*}{\cmark}&\multirow{1}{*}{$3\slash 3$} \\

   Pear & \multirow{1}{*}{\cmark}&\multirow{1}{*}{\cmark}&\multirow{1}{*}{\cmark}&\multirow{1}{*}{$3\slash 3$}&\multirow{1}{*}{-}&\multirow{1}{*}{-}&\multirow{1}{*}{-}&\multirow{1}{*}{-} \\

   Strawberry & \multirow{1}{*}{\cmark}&\multirow{1}{*}{\cmark}&\multirow{1}{*}{\cmark}&\multirow{1}{*}{$3\slash 3$}&\multirow{1}{*}{-}&\multirow{1}{*}{-}&\multirow{1}{*}{-}&\multirow{1}{*}{-} \\

   Small Clamp & \multirow{1}{*}{\cmark}&\multirow{1}{*}{\cmark}&\multirow{1}{*}{\cmark}&\multirow{1}{*}{$3\slash 3$}&\multirow{1}{*}{-}&\multirow{1}{*}{-}&\multirow{1}{*}{-}&\multirow{1}{*}{-} \\

   Medium Clamp & \multirow{1}{*}{\cmark}&\multirow{1}{*}{\cmark}&\multirow{1}{*}{\cmark}&\multirow{1}{*}{$3\slash 3$}&\multirow{1}{*}{-}&\multirow{1}{*}{-}&\multirow{1}{*}{-}&\multirow{1}{*}{-} \\

   Large Clamp & \multirow{1}{*}{\cmark}&\multirow{1}{*}{\cmark}&\multirow{1}{*}{\cmark}&\multirow{1}{*}{$3\slash 3$}&\multirow{1}{*}{-}&\multirow{1}{*}{-}&\multirow{1}{*}{-}&\multirow{1}{*}{-} \\

   \textbf{Overall} & & & & \textbf{$95.6\%$} & & & & \textbf{$90\%$}\\
 \bottomrule
\end{tabular}}
\caption{Detailed grasp stability results involving various household items from the YCB dataset. Most of our grasp are stable enough for object handover. When the part is specified by the user, the stability drops marginally as the grasping position is more constrained.} \label{tab:grasp_results}
\vspace{-0.1in}
\end{table}

The detailed testing outcomes are listed in Table \ref{tab:grasp_results}. Overall, the robot grasping pose generated by our multimodal approach is stable enough for handover in more than $90 \%$ of cases. The stability rate drops slightly when parts are specified as they introduces more constraints. The failure cases mostly involved objects (``Scissors'', ``Screwdriver'', ``Small Clamp'') that were narrow and close to the tabletop. The small size reduced the physical region where stable grasps can be found, and degraded the pointcloud, which reduced the number of grasp points visible to the model. These factors combined increased the difficulty of successfully grasping small, stout, objects.  

\begin{table}[htbp]
\centering
\scalebox{0.8}{
\label{tab:execution_time}
\begin{tabular}{lr}\toprule
\textbf{Process} & \textbf{Time} \\ \midrule
Display scene to user & Instant \\
User utters request & Varies by user \\
Object selection & 4 seconds \\
\hspace{0.3cm}$\hookrightarrow$ With part selection & 7 seconds \\
Robot grasp prediction & 17 seconds \\
\hspace{0.3cm}$\hookrightarrow$ With human grasp prediction & 32 seconds \\
Arm picks up target object & 5 seconds \\ \vspace{0.2cm}
Perform handover to human & 14 seconds \\ 
\textbf{Total with only object specification} & 55 seconds \\
\textbf{Total with part and object specification} & 43 seconds \\
\bottomrule
\end{tabular}}
\caption{Standard execution time for each module of our pipeline during a full run. User feedback for time consumption may be influenced by the long processing time spent on the robot and human grasp prediction. Note that we report the time upper bound for the human grasp prediction step because it involves stochasticity.} 
\label{tab:breakdown}
\vspace{0.2in}
\vspace{-0.2in}
\end{table}
\subsection{User Study}
\label{subsec:user_study}
To comprehensively evaluate the entire multimodal pipe-line for precise object handover, we conducted a user study involving 10 randomly chosen participants with no prior knowledge about the system. Fig. \ref{fig:real_exp_figure_1} and Fig. \ref{fig:real_exp_figure_2} depicts the sample workflow in our study. We first give each user a test run to get them familiar with the objective of our approach as well as how to give out gaze and language instructions. When the formal test begins, each of the users is presented with three scenes from the monitor comprising random objects sourced from the YCB dataset. For each environment, they are asked to select two objects of interest in order for handover. When the scene is projected on the user's monitor, we also display the current gaze detection result on top of the image in real-time. This move helps the participants verify if they are focusing on the correct object and adjust their eye focal point if needed. When both gaze and verbal inputs are captured, the rest of the pipeline is executed, culminating in the handover of the objects to the user by the robot arm. After all the trials, the users are asked to fill in a questionnaire composed of the following questions:
\begin{itemize}
    \item Does the system select the desired object for handover?
    \item Does the robot successfully grasp the desired part of the object?
    \item Does the robot successfully handover the desired object? 
    \item Does the robot leave enough margin for you to grasp the object?
    \item Is the system's time consumption to pick and handover object accpetable?
\end{itemize}

These questions focuses on critical subjective factors of our approach including the grasp suitability, the efficiency of handover in terms of comfort and speed, and the promotion of safe human-robot interaction. This methodology allowed us to gain insights into the nuanced aspects of human-robot interaction, shedding light on the effectiveness and user experience of our system beyond mere quantitative measures. In addition, the users are also welcomed to leave any additional comments.
%\textbf{Handover Effectiveness:} To comprehensively evaluate the subjective aspects of the grasp subsystem and human interaction, we conducted a user study alongside objective metrics such as object/grasp selection accuracy. Our study involved 10 users, each presented with three scenes comprising random objects sourced from the YCB dataset \cite{YCBDataset}. In each scenario, users were presented with an image of the scene and instructed to issue two commands while looking at the objects of interest. Upon the conclusion of the language input, the entire pipeline is executed, culminating in the handover of the objects (each object done seperately) to the user. Our evaluation focused on critical subjective factors including grasp suitability, the efficiency of handover in terms of comfort and speed, and the promotion of safe human-robot interaction. This methodology allowed us to gain insights into the nuanced aspects of human-robot interaction, shedding light on the effectiveness and user experience of our system beyond mere quantitative measures.

\iffalse
\begin{figure}[htbp]
\centering
\includegraphics[width = 5cm]{images/usrstudy.png}
\caption{A survey including the above questions is taken after the user study. The above image conveys the average score per question across users for each scene. Furthermore, the overall average bar finds the mean across users and scenes per question. Most users have positive experiences with our multimodal approach for precise object handover.}
\label{usrstudyresults}
\end{figure}
\fi

\begin{figure}[h]
    \vspace{-10pt}
    \centering
    \includegraphics[trim = {4cm 0cm 0cm 1cm}, clip, width = 0.6\textwidth]{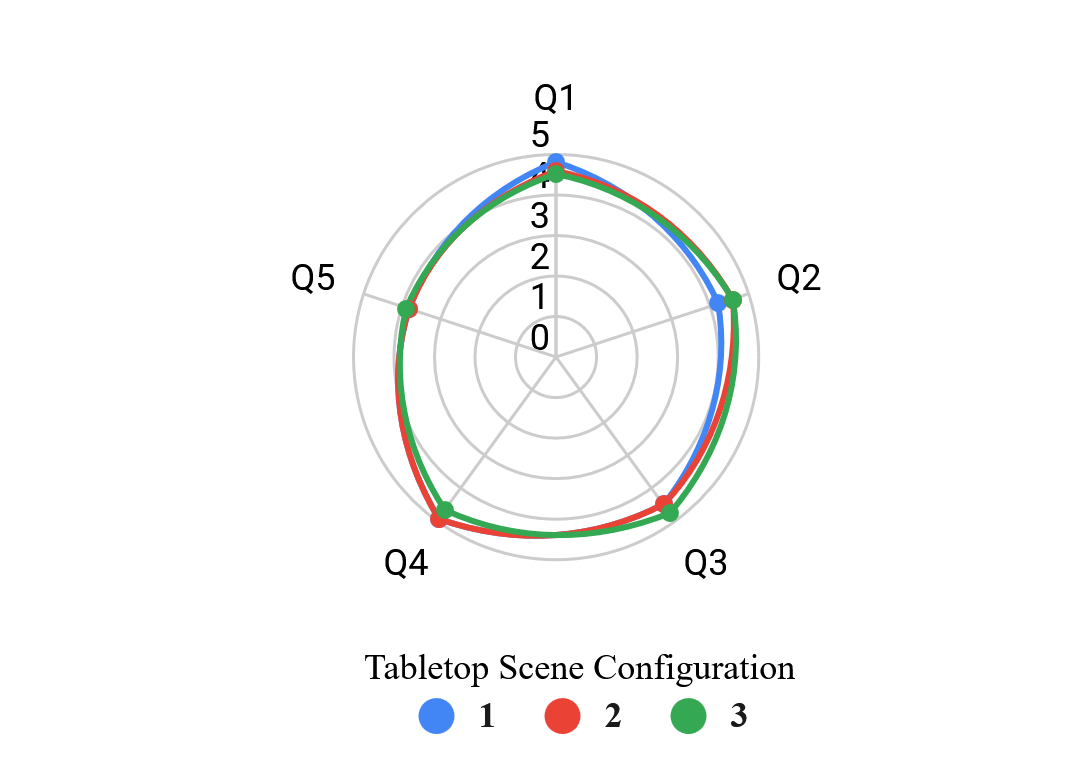}
    \caption{A survey including the above questions is taken after the user study. The above image conveys the average score per question across users for each scene. Most users had a positive experience with our multimodal approach for precise object handover, giving average ratings above 4, with the exception of question 5 regarding time consumption.}
    \label{fig:user_study_res}
\end{figure}

The plot regarding the user study results is shown in Fig. \ref{fig:user_study_res}. In the study, our multi-modal approach accurately extracts the objects of interest and successfully hands them over to the users most of the time. Many participants stated that the handover process was smooth as the robot left enough safety margin for them to retrieve the object easily. The only question with an average score below 4 is about the system's time consumption. The execution time of the whole pipeline normally takes around 50 seconds from the initial scene capture to the final handover. Since all users treat our system as a black box without knowing the complex architecture and the heavy computation involved, they may have unrealistic expectations, such as expecting the robot to get the object for them immediately after the instructions are given. In Table \ref{tab:breakdown}, we provide a detailed breakdown of the time contribution for each module in our system, and find that a majority of the time is spent on human hand prediction and robot grasp generation. The human grasp prediction involves a random sampling phase, which consumes a good amount of time to find a viable and stable grasp. Additionally, in the robot grasp generation module, the runtime is quadratic as we need to iterate for every unique robot and human grasp pair.  In future work, we plan to speed up the whole process by parallelizing the grasp generation step, as its computational nature allows for the operation to be split across multiple threads.

\section{Conclusions and Future Work}
In this paper, we present a system that uses multi-modal human inputs and prior knowledge of grasping preferences to perform accurate and comfortable object handover. While past works focus on handover within a single modality, we integrated gaze information, fine-grained verbal instructions, and grasping preferences to achieve a more accurate understanding of the user's intention. Modular experiments on our pipeline show how each modality contributes towards improving object and part selection accuracy, grasping stability, and human-aware grasp selection. A user study conducted in a real setting yielded favorable user feedback, notably for selection accuracy and handover comfort. In future works, we aim to improve pipeline execution time as well as increase the richness of verbal preferences supported by the system.

\section*{Funding}
This work was supported by the National Science Foundation (NSF) under award no. 2204528.

\section*{Ethics approval}
Our human subject study design was approved by Purdue University IRB under app. IRB-2024-1687.

\section*{CRediT authorship contribution statement}
\textbf{Lucas Chen}: Conceptualization, Methodology, Software, Validation, Writing-Original Draft. \textbf{Guna Avula}: Conceptualization, Methodology, Software, Validation, Wr-iting-Original Draft. \textbf{Hanwen Ren}: Conceptualization, Formal analysis, Visualization, Writing-review and Editing. \textbf{Zixing Wang}: Conceptualization, Writing-Review and Editing. \textbf{Ahmed H. Qureshi}: Funding acquisition, Project administration, and Supervision.

\section*{Declaration of competing interest}
The authors declare that they have no known competing financial interests or personal relationships that could have appeared to influence the work reported in this paper.

\section*{Data availability}
The data will be made available on request.

\bibliographystyle{elsarticle-num} 
\bibliography{sn-bibliography}

\begin{thebibliography}{10}
\expandafter\ifx\csname url\endcsname\relax
  \def\url#1{\texttt{#1}}\fi
\expandafter\ifx\csname urlprefix\endcsname\relax\def\urlprefix{URL }\fi
\expandafter\ifx\csname href\endcsname\relax
  \def\href#1#2{#2} \def\path#1{#1}\fi

\bibitem{ortenzi2022object}
V.~Ortenzi, A.~Cosgun, T.~Pardi, W.~Chan, E.~Croft, D.~Kulic, Object handovers: a review for robotics, IEEE Transactions on robotics (2022).
\newblock \href {http://arxiv.org/abs/2007.12952} {\path{arXiv:2007.12952}}.

\bibitem{DUAN2024100145}
H.~Duan, Y.~Yang, D.~Li, P.~Wang, \href{https://www.sciencedirect.com/science/article/pii/S2667379724000032}{Human–robot object handover: Recent progress and future direction}, Biomimetic Intelligence and Robotics 4~(1) (2024) 100145.
\newblock \href {https://doi.org/https://doi.org/10.1016/j.birob.2024.100145} {\path{doi:https://doi.org/10.1016/j.birob.2024.100145}}.
\newline\urlprefix\url{https://www.sciencedirect.com/science/article/pii/S2667379724000032}

\bibitem{Aleotti2014AnAS}
J.~Aleotti, V.~Micelli, S.~Caselli, \href{https://api.semanticscholar.org/CorpusID:45666433}{An affordance sensitive system for robot to human object handover}, International Journal of Social Robotics 6 (2014) 653 -- 666.
\newline\urlprefix\url{https://api.semanticscholar.org/CorpusID:45666433}

\bibitem{Dang2012SemanticGP}
H.~N. Dang, P.~K. Allen, \href{https://api.semanticscholar.org/CorpusID:8860232}{Semantic grasping: Planning robotic grasps functionally suitable for an object manipulation task}, 2012 IEEE/RSJ International Conference on Intelligent Robots and Systems (2012) 1311--1317.
\newline\urlprefix\url{https://api.semanticscholar.org/CorpusID:8860232}

\bibitem{5326233}
M.~Huber, H.~Radrich, C.~Wendt, M.~Rickert, A.~Knoll, T.~Brandt, S.~Glasauer, Evaluation of a novel biologically inspired trajectory generator in human-robot interaction, in: RO-MAN 2009 - The 18th IEEE International Symposium on Robot and Human Interactive Communication, 2009, pp. 639--644.
\newblock \href {https://doi.org/10.1109/ROMAN.2009.5326233} {\path{doi:10.1109/ROMAN.2009.5326233}}.

\bibitem{liu2024hoi4d}
Y.~Liu, Y.~Liu, C.~Jiang, K.~Lyu, W.~Wan, H.~Shen, B.~Liang, Z.~Fu, H.~Wang, L.~Yi, Hoi4d: A 4d egocentric dataset for category-level human-object interaction, Computer Vision and Pattern Recognition (2024).
\newblock \href {http://arxiv.org/abs/2203.01577} {\path{arXiv:2203.01577}}.

\bibitem{wei2024grasp}
Y.-L. Wei, J.-J. Jiang, C.~Xing, X.~Tan, X.-M. Wu, H.~Li, M.~Cutkosky, W.-S. Zheng, Grasp as you say: Language-guided dexterous grasp generation (2024).
\newblock \href {http://arxiv.org/abs/2405.19291} {\path{arXiv:2405.19291}}.

\bibitem{Kim1992HandoverOA}
I.~Kim, H.~Inooka, \href{https://api.semanticscholar.org/CorpusID:62094396}{Hand-over of an object between human and robot}, [1992] Proceedings IEEE International Workshop on Robot and Human Communication (1992) 199--203.
\newline\urlprefix\url{https://api.semanticscholar.org/CorpusID:62094396}

\bibitem{10160623}
A.~K. Keshari, H.~Ren, A.~H. Qureshi, Cograsp: 6-dof grasp generation for human-robot collaboration, in: 2023 IEEE International Conference on Robotics and Automation (ICRA), 2023, pp. 9829--9836.
\newblock \href {https://doi.org/10.1109/ICRA48891.2023.10160623} {\path{doi:10.1109/ICRA48891.2023.10160623}}.

\bibitem{8206205}
S.~Parastegari, B.~Abbasi, E.~Noohi, M.~Zefran, Modeling human reaching phase in human-human object handover with application in robot-human handover, in: 2017 IEEE/RSJ International Conference on Intelligent Robots and Systems (IROS), 2017, pp. 3597--3602.
\newblock \href {https://doi.org/10.1109/IROS.2017.8206205} {\path{doi:10.1109/IROS.2017.8206205}}.

\bibitem{8341961}
S.~Parastegari, E.~Noohi, B.~Abbasi, M.~Žefran, Failure recovery in robot–human object handover, IEEE Transactions on Robotics 34~(3) (2018) 660--673.
\newblock \href {https://doi.org/10.1109/TRO.2018.2819198} {\path{doi:10.1109/TRO.2018.2819198}}.

\bibitem{eeg_cursor}
J.~R. Wolpaw, D.~J. McFarland, G.~W. Neat, C.~A. Forneris, \href{https://www.sciencedirect.com/science/article/pii/001346949190040B}{An eeg-based brain-computer interface for cursor control}, Electroencephalography and Clinical Neurophysiology 78~(3) (1991) 252--259.
\newblock \href {https://doi.org/https://doi.org/10.1016/0013-4694(91)90040-B} {\path{doi:https://doi.org/10.1016/0013-4694(91)90040-B}}.
\newline\urlprefix\url{https://www.sciencedirect.com/science/article/pii/001346949190040B}

\bibitem{eeg_robot_arm}
J.-H. Jeong, K.-H. Shim, D.-J. Kim, S.-W. Lee, Brain-controlled robotic arm system based on multi-directional cnn-bilstm network using eeg signals, IEEE Transactions on Neural Systems and Rehabilitation Engineering 28~(5) (2020) 1226--1238.
\newblock \href {https://doi.org/10.1109/TNSRE.2020.2981659} {\path{doi:10.1109/TNSRE.2020.2981659}}.

\bibitem{devlin2019bert}
J.~Devlin, M.-W. Chang, K.~Lee, K.~Toutanova, Bert: Pre-training of deep bidirectional transformers for language understanding, North American Chapter of the Association for Computational Linguistics (2019).
\newblock \href {http://arxiv.org/abs/1810.04805} {\path{arXiv:1810.04805}}.

\bibitem{raffel2023exploring}
C.~Raffel, N.~Shazeer, A.~Roberts, K.~Lee, S.~Narang, M.~Matena, Y.~Zhou, W.~Li, P.~J. Liu, Exploring the limits of transfer learning with a unified text-to-text transformer, Journal of machine learning research (2023).
\newblock \href {http://arxiv.org/abs/1910.10683} {\path{arXiv:1910.10683}}.

\bibitem{clark2020electra}
K.~Clark, M.-T. Luong, Q.~V. Le, C.~D. Manning, Electra: Pre-training text encoders as discriminators rather than generators, International Conference on Learning Representations (ICLR) (2020).
\newblock \href {http://arxiv.org/abs/2003.10555} {\path{arXiv:2003.10555}}.

\bibitem{brown2020language}
T.~B. Brown, B.~Mann, N.~Ryder, M.~Subbiah, J.~Kaplan, P.~Dhariwal, A.~Neelakantan, P.~Shyam, G.~Sastry, A.~Askell, S.~Agarwal, A.~Herbert-Voss, G.~Krueger, T.~Henighan, R.~Child, A.~Ramesh, D.~M. Ziegler, J.~Wu, C.~Winter, C.~Hesse, M.~Chen, E.~Sigler, M.~Litwin, S.~Gray, B.~Chess, J.~Clark, C.~Berner, S.~McCandlish, A.~Radford, I.~Sutskever, D.~Amodei, Language models are few-shot learners, Neural Information Processing Systems (2020).
\newblock \href {http://arxiv.org/abs/2005.14165} {\path{arXiv:2005.14165}}.

\bibitem{LISA}
X.~Lai, Z.~Tian, Y.~Chen, Y.~Li, Y.~Yuan, S.~Liu, J.~Jia, \href{https://api.semanticscholar.org/CorpusID:260351258}{Lisa: Reasoning segmentation via large language model}, 2024 IEEE/CVF Conference on Computer Vision and Pattern Recognition (CVPR) (2023) 9579--9589.
\newline\urlprefix\url{https://api.semanticscholar.org/CorpusID:260351258}

\bibitem{3d_llm_segmentation_1}
R.~Huang, X.~Pan, H.~Zheng, H.~Jiang, Z.~Xie, S.~Song, G.~Huang, Joint representation learning for text and 3d point cloud, Pattern Recognition (2023).
\newblock \href {http://arxiv.org/abs/2301.07584} {\path{arXiv:2301.07584}}.

\bibitem{3d_llm_segmentation_2}
D.~Rozenberszki, O.~Litany, A.~Dai, Language-grounded indoor 3d semantic segmentation in the wild, European Conference on Computer Vision (2022).
\newblock \href {http://arxiv.org/abs/2204.07761} {\path{arXiv:2204.07761}}.

\bibitem{3d_llm_segmentation_3}
Z.~L. Lin, X.~Peng, P.~Cong, Y.~Hou, X.~Zhu, S.~Yang, Y.~Ma, \href{https://api.semanticscholar.org/CorpusID:258079297}{Wildrefer: 3d object localization in large-scale dynamic scenes with multi-modal visual data and natural language}, in: European Conference on Computer Vision, 2023.
\newline\urlprefix\url{https://api.semanticscholar.org/CorpusID:258079297}

\bibitem{gaze_survey}
S.~Ghosh, A.~Dhall, M.~Hayat, J.~Knibbe, Q.~Ji, Automatic gaze analysis: A survey of deep learning based approaches, IEEE Transactions on Pattern Analysis and Machine Intelligence (2022).
\newblock \href {http://arxiv.org/abs/2108.05479} {\path{arXiv:2108.05479}}.

\bibitem{zhang2017mpiigaze}
X.~Zhang, Y.~Sugano, M.~Fritz, A.~Bulling, Mpiigaze: Real-world dataset and deep appearance-based gaze estimation, IEEE Transactions on Pattern Analysis and Machine Intelligence (2017).
\newblock \href {http://arxiv.org/abs/1711.09017} {\path{arXiv:1711.09017}}.

\bibitem{zhang19_pami}
X.~Zhang, Y.~Sugano, M.~Fritz, A.~Bulling, Mpiigaze: Real-world dataset and deep appearance-based gaze estimation, IEEE Transactions on Pattern Analysis and Machine Intelligence (TPAMI) 41~(1) (2019) 162--175.
\newblock \href {https://doi.org/10.1109/TPAMI.2017.2778103} {\path{doi:10.1109/TPAMI.2017.2778103}}.

\bibitem{ccakir2023reviewing}
M.~{\c{C}}ak{\i}r, A.~Huckauf, Reviewing the social function of eye gaze in social interaction, in: Proceedings of the 2023 Symposium on Eye Tracking Research and Applications, 2023, pp. 1--3.

\bibitem{Strabala2013TowardSH}
K.~Strabala, M.~K. Lee, A.~D. Dragan, J.~Forlizzi, S.~S. Srinivasa, M.~Cakmak, V.~Micelli, \href{https://api.semanticscholar.org/CorpusID:15247009}{Toward seamless human-robot handovers}, Journal of Human-Robot Interaction 2 (2013) 112 -- 132.
\newline\urlprefix\url{https://api.semanticscholar.org/CorpusID:15247009}

\bibitem{qiao2020referring}
Y.~Qiao, C.~Deng, Q.~Wu, Referring expression comprehension: A survey of methods and datasets, IEEE transactions on multimedia (2020).
\newblock \href {http://arxiv.org/abs/2007.09554} {\path{arXiv:2007.09554}}.

\bibitem{Langer2022ILG}
D.~Langer, F.~Legler, P.~Kotsch, A.~Dettmann, A.~C. Bullinger-Hoffmann, \href{https://api.semanticscholar.org/CorpusID:252973599}{I let go now! towards a voice-user interface for handovers between robots and users with full and impaired sight}, Robotics 11 (2022) 112.
\newline\urlprefix\url{https://api.semanticscholar.org/CorpusID:252973599}

\bibitem{zhang2024invigorate}
H.~Zhang, Y.~Lu, C.~Yu, D.~Hsu, X.~Lan, N.~Zheng, Invigorate: Interactive visual grounding and grasping in clutter, Robotics: Science and Systems (2024).
\newblock \href {http://arxiv.org/abs/2108.11092} {\path{arXiv:2108.11092}}.

\bibitem{chen2022gscorecam}
P.~Chen, Q.~Li, S.~Biaz, T.~Bui, A.~Nguyen, gscorecam: What objects is clip looking at?, in: Proceedings of the Asian Conference on Computer Vision, 2022, pp. 1959--1975.

\bibitem{lueddecke22_cvpr}
T.~L\"uddecke, A.~Ecker, Image segmentation using text and image prompts, in: Proceedings of the IEEE/CVF Conference on Computer Vision and Pattern Recognition (CVPR), 2022, pp. 7086--7096.

\bibitem{Decatur20223DHL}
D.~Decatur, I.~Lang, R.~Hanocka, \href{https://api.semanticscholar.org/CorpusID:254926514}{3d highlighter: Localizing regions on 3d shapes via text descriptions}, 2023 IEEE/CVF Conference on Computer Vision and Pattern Recognition (CVPR) (2022) 20930--20939.
\newline\urlprefix\url{https://api.semanticscholar.org/CorpusID:254926514}

\bibitem{song2023learning}
Y.~Song, P.~Sun, Y.~Ren, Y.~Zheng, Y.~Zhang, Learning 6-dof fine-grained grasp detection based on part affordance grounding, CoRR (2023).
\newblock \href {http://arxiv.org/abs/2301.11564} {\path{arXiv:2301.11564}}.

\bibitem{Zhu2022PointCLIPVP}
X.~Zhu, R.~Zhang, B.~He, Z.~Guo, Z.~Zeng, Z.~Qin, S.~Zhang, P.~Gao, \href{https://api.semanticscholar.org/CorpusID:261241594}{Pointclip v2: Prompting clip and gpt for powerful 3d open-world learning}, 2023 IEEE/CVF International Conference on Computer Vision (ICCV) (2022) 2639--2650.
\newline\urlprefix\url{https://api.semanticscholar.org/CorpusID:261241594}

\bibitem{Ngyen2023OpenVocabularyAD}
T.~Ngyen, M.~N. Vu, A.~Vuong, D.~Nguyen, T.~D. Vo, N.~T.~H. Le, A.~M. Nguyen, \href{https://api.semanticscholar.org/CorpusID:257365163}{Open-vocabulary affordance detection in 3d point clouds}, 2023 IEEE/RSJ International Conference on Intelligent Robots and Systems (IROS) (2023) 5692--5698.
\newline\urlprefix\url{https://api.semanticscholar.org/CorpusID:257365163}

\bibitem{Tang2023TaskOrientedGP}
C.~Tang, D.~Huang, L.~Meng, W.~Liu, H.~Zhang, \href{https://api.semanticscholar.org/CorpusID:257233075}{Task-oriented grasp prediction with visual-language inputs}, 2023 IEEE/RSJ International Conference on Intelligent Robots and Systems (IROS) (2023) 4881--4888.
\newline\urlprefix\url{https://api.semanticscholar.org/CorpusID:257233075}

\bibitem{Ding2020PhraseClickTA}
H.~Ding, S.~D. Cohen, B.~L. Price, X.~Jiang, \href{https://api.semanticscholar.org/CorpusID:224801564}{Phraseclick: Toward achieving flexible interactive segmentation by phrase and click}, in: European Conference on Computer Vision, 2020.
\newline\urlprefix\url{https://api.semanticscholar.org/CorpusID:224801564}

\bibitem{Leal2020INITIATINGOH}
D.~Leal, O.~Leal, \href{https://api.semanticscholar.org/CorpusID:231853694}{Initiating object handover in human-robot collaboration using a multi-modal wearable and deep learning visual system}, 2020.
\newline\urlprefix\url{https://api.semanticscholar.org/CorpusID:231853694}

\bibitem{9515402}
J.~Laplaza, A.~Pumarola, F.~Moreno-Noguer, A.~Sanfeliu, Attention deep learning based model for predicting the 3d human body pose using the robot human handover phases, in: 2021 30th IEEE International Conference on Robot and Human Interactive Communication (RO-MAN), 2021, pp. 161--166.
\newblock \href {https://doi.org/10.1109/RO-MAN50785.2021.9515402} {\path{doi:10.1109/RO-MAN50785.2021.9515402}}.

\bibitem{choi2022preemptive}
A.~Choi, M.~K. Jawed, J.~Joo, Preemptive motion planning for human-to-robot indirect placement handovers, in: 2022 International Conference on Robotics and Automation (ICRA), IEEE, 2022, pp. 4743--4749.

\bibitem{Cosgun2015DidYM}
A.~Cosgun, A.~J.~B. Trevor, H.~I. Christensen, \href{https://api.semanticscholar.org/CorpusID:211567531}{Did you mean this object?: Detecting ambiguity in pointing gesture targets}, 2015.
\newline\urlprefix\url{https://api.semanticscholar.org/CorpusID:211567531}

\bibitem{trick2019multimodal}
S.~Trick, D.~Koert, J.~Peters, C.~A. Rothkopf, Multimodal uncertainty reduction for intention recognition in human-robot interaction, in: 2019 IEEE/RSJ International Conference on Intelligent Robots and Systems (IROS), IEEE, 2019, pp. 7009--7016.

\bibitem{rajabi2023detecting}
N.~Rajabi, P.~Khanna, S.~U.~D. Kanik, E.~Yadollahi, M.~Vasco, M.~Bj{\"o}rkman, C.~Smith, D.~Kragic, Detecting the intention of object handover in human-robot collaborations: An eeg study, in: 2023 32nd IEEE International Conference on Robot and Human Interactive Communication (RO-MAN), IEEE, 2023, pp. 549--555.

\bibitem{sharma2022towards}
M.~Sharma, M.~Rekrut, J.~Alexandersson, A.~Kr{\"u}ger, Towards improving eeg-based intent recognition in visual search tasks, in: International Conference on Neural Information Processing, Springer, 2022, pp. 604--615.

\bibitem{cooper2020eeg}
S.~Cooper, S.~F. Fensome, D.~Kourtis, S.~Gow, M.~Dragone, An eeg investigation on planning human-robot handover tasks, in: 2020 IEEE International Conference on Human-Machine Systems (ICHMS), IEEE, 2020, pp. 1--6.

\bibitem{Vasudevan2018ObjectRI}
A.~B. Vasudevan, D.~Dai, L.~V. Gool, \href{https://api.semanticscholar.org/CorpusID:4576781}{Object referring in videos with language and human gaze}, 2018 IEEE/CVF Conference on Computer Vision and Pattern Recognition (2018) 4129--4138.
\newline\urlprefix\url{https://api.semanticscholar.org/CorpusID:4576781}

\bibitem{Tian2024GazeguidedHI}
J.~Tian, L.~Yang, R.~Ji, Y.~Ma, L.~Xu, J.~Yu, Y.~Shi, J.~Wang, \href{https://api.semanticscholar.org/CorpusID:268681107}{Gaze-guided hand-object interaction synthesis: Benchmark and method}, ArXiv abs/2403.16169 (2024).
\newline\urlprefix\url{https://api.semanticscholar.org/CorpusID:268681107}

\bibitem{staudte2008utility}
M.~Staudte, M.~Crocker, The utility of gaze in spoken human-robot interaction, in: Proceedings of Workshop on Metrics for Human-Robot Interaction 2008, March 12th, 2008, pp. 53--59.

\bibitem{Moon2014MeetMW}
A.~Moon, D.~M. Troniak, B.~T. Gleeson, M.~K. X.~J. Pan, M.~Zheng, B.~A. Blumer, K.~E. Maclean, E.~A. Croft, \href{https://api.semanticscholar.org/CorpusID:13576005}{Meet me where i’m gazing: How shared attention gaze affects human-robot handover timing}, 2014 9th ACM/IEEE International Conference on Human-Robot Interaction (HRI) (2014) 334--341.
\newline\urlprefix\url{https://api.semanticscholar.org/CorpusID:13576005}

\bibitem{understandtranfer}
M.~Gharbi, P.-V. Paubel, A.~Clodic, O.~Carreras, R.~Alami, J.-M. Cellier, Toward a better understanding of the communication cues involved in a human-robot object transfer, in: 2015 24th IEEE International Symposium on Robot and Human Interactive Communication (RO-MAN), 2015, pp. 319--324.
\newblock \href {https://doi.org/10.1109/ROMAN.2015.7333626} {\path{doi:10.1109/ROMAN.2015.7333626}}.

\bibitem{Aronson2018EyeHandBI}
R.~M. Aronson, T.~Santini, T.~C. K{\"u}bler, E.~Kasneci, S.~S. Srinivasa, H.~Admoni, \href{https://api.semanticscholar.org/CorpusID:3707577}{Eye-hand behavior in human-robot shared manipulation}, 2018 13th ACM/IEEE International Conference on Human-Robot Interaction (HRI) (2018) 4--13.
\newline\urlprefix\url{https://api.semanticscholar.org/CorpusID:3707577}

\bibitem{dwellpaulus}
Y.~T. Paulus, G.~B. Remijn, \href{https://www.sciencedirect.com/science/article/pii/S0141938221000123}{Usability of various dwell times for eye-gaze-based object selection with eye tracking}, Displays 67 (2021) 101997.
\newblock \href {https://doi.org/https://doi.org/10.1016/j.displa.2021.101997} {\path{doi:https://doi.org/10.1016/j.displa.2021.101997}}.
\newline\urlprefix\url{https://www.sciencedirect.com/science/article/pii/S0141938221000123}

\bibitem{eyepointing}
R.~Schweigert, V.~Schwind, S.~Mayer, \href{https://doi.org/10.1145/3340764.3344897}{Eyepointing: A gaze-based selection technique}, in: Proceedings of Mensch Und Computer 2019, MuC '19, Association for Computing Machinery, New York, NY, USA, 2019, p. 719–723.
\newblock \href {https://doi.org/10.1145/3340764.3344897} {\path{doi:10.1145/3340764.3344897}}.
\newline\urlprefix\url{https://doi.org/10.1145/3340764.3344897}

\bibitem{shao2015eyelasso}
Y.-F. Shao, C.~Wang, C.-S. Fuh, Eyelasso: Real-world object selection using gaze-based gestures, in: 28th IPPR Conference on Computer Vision, Graphics, and Image Processing, 2015.

\bibitem{krafka2016eye}
K.~Krafka, A.~Khosla, P.~Kellnhofer, H.~Kannan, S.~Bhandarkar, W.~Matusik, A.~Torralba, Eye tracking for everyone, Computer Vision and Pattern Recognition (2016).
\newblock \href {http://arxiv.org/abs/1606.05814} {\path{arXiv:1606.05814}}.

\bibitem{Deng2017MonocularF3}
H.~Deng, W.~Zhu, \href{https://api.semanticscholar.org/CorpusID:32485021}{Monocular free-head 3d gaze tracking with deep learning and geometry constraints}, 2017 IEEE International Conference on Computer Vision (ICCV) (2017) 3162--3171.
\newline\urlprefix\url{https://api.semanticscholar.org/CorpusID:32485021}

\bibitem{FischerECCV2018}
T.~Fischer, H.~J. Chang, Y.~Demiris, {RT-GENE: Real-Time Eye Gaze Estimation in Natural Environments}, in: European Conference on Computer Vision, 2018, pp. 339--357.

\bibitem{kellnhofer2019gaze360}
P.~Kellnhofer, A.~Recasens, S.~Stent, W.~Matusik, A.~Torralba, Gaze360: Physically unconstrained gaze estimation in the wild, IEEE International Conference on Computer Vision (2019).
\newblock \href {http://arxiv.org/abs/1910.10088} {\path{arXiv:1910.10088}}.

\bibitem{Smith2013GazeLP}
B.~A. Smith, Q.~Yin, S.~K. Feiner, S.~K. Nayar, \href{https://api.semanticscholar.org/CorpusID:16104238}{Gaze locking: passive eye contact detection for human-object interaction}, Proceedings of the 26th annual ACM symposium on User interface software and technology (2013).
\newline\urlprefix\url{https://api.semanticscholar.org/CorpusID:16104238}

\bibitem{McMurrough2012AnET}
C.~D. McMurrough, V.~Metsis, J.~Rich, F.~Makedon, \href{https://api.semanticscholar.org/CorpusID:466361}{An eye tracking dataset for point of gaze detection}, Proceedings of the Symposium on Eye Tracking Research and Applications (2012).
\newline\urlprefix\url{https://api.semanticscholar.org/CorpusID:466361}

\bibitem{Mora2014EYEDIAPAD}
K.~A.~F. Mora, F.~Monay, J.-M. Odobez, \href{https://api.semanticscholar.org/CorpusID:396361}{Eyediap: a database for the development and evaluation of gaze estimation algorithms from rgb and rgb-d cameras}, Proceedings of the Symposium on Eye Tracking Research and Applications (2014).
\newline\urlprefix\url{https://api.semanticscholar.org/CorpusID:396361}

\bibitem{4449972}
U.~Weidenbacher, G.~Layher, P.-M. Strauss, H.~Neumann, A comprehensive head pose and gaze database, in: 2007 3rd IET International Conference on Intelligent Environments, 2007, pp. 455--458.
\newblock \href {https://doi.org/10.1049/cp:20070407} {\path{doi:10.1049/cp:20070407}}.

\bibitem{Hennessey2006ASC}
C.~Hennessey, B.~Noureddin, P.~D. Lawrence, \href{https://api.semanticscholar.org/CorpusID:3238368}{A single camera eye-gaze tracking system with free head motion}, Proceedings of the 2006 symposium on Eye tracking research \& applications (2006).
\newline\urlprefix\url{https://api.semanticscholar.org/CorpusID:3238368}

\bibitem{Pan2017AutomatedDO}
M.~K. X.~J. Pan, V.~Skjerv{\o}y, W.~P. Chan, M.~Inaba, E.~A. Croft, \href{https://api.semanticscholar.org/CorpusID:43614132}{Automated detection of handovers using kinematic features}, The International Journal of Robotics Research 36 (2017) 721 -- 738.
\newline\urlprefix\url{https://api.semanticscholar.org/CorpusID:43614132}

\bibitem{Micelli2011PerceptionAC}
V.~Micelli, K.~Strabala, S.~S. Srinivasa, \href{https://api.semanticscholar.org/CorpusID:15607569}{Perception and control challenges for effec tive human-robot handoff s}, 2011.
\newline\urlprefix\url{https://api.semanticscholar.org/CorpusID:15607569}

\bibitem{8579107}
W.~Wang, R.~Li, Z.~M. Diekel, Y.~Chen, Z.~Zhang, Y.~Jia, Controlling object hand-over in human–robot collaboration via natural wearable sensing, IEEE Transactions on Human-Machine Systems 49~(1) (2019) 59--71.
\newblock \href {https://doi.org/10.1109/THMS.2018.2883176} {\path{doi:10.1109/THMS.2018.2883176}}.

\bibitem{6281385}
M.~Cakmak, S.~S. Srinivasa, M.~K. Lee, S.~Kiesler, J.~Forlizzi, Using spatial and temporal contrast for fluent robot-human hand-overs, in: 2011 6th ACM/IEEE International Conference on Human-Robot Interaction (HRI), 2011, pp. 489--496.
\newblock \href {https://doi.org/10.1145/1957656.1957823} {\path{doi:10.1145/1957656.1957823}}.

\bibitem{9341004}
W.~Yang, C.~Paxton, M.~Cakmak, D.~Fox, Human grasp classification for reactive human-to-robot handovers, in: 2020 IEEE/RSJ International Conference on Intelligent Robots and Systems (IROS), 2020, pp. 11123--11130.
\newblock \href {https://doi.org/10.1109/IROS45743.2020.9341004} {\path{doi:10.1109/IROS45743.2020.9341004}}.

\bibitem{yang2021reactive}
W.~Yang, C.~Paxton, A.~Mousavian, Y.-W. Chao, M.~Cakmak, D.~Fox, Reactive human-to-robot handovers of arbitrary objects (2021).
\newblock \href {http://arxiv.org/abs/2011.08961} {\path{arXiv:2011.08961}}.

\bibitem{6343845}
J.~Aleotti, V.~Micelli, S.~Caselli, Comfortable robot to human object hand-over, in: 2012 IEEE RO-MAN: The 21st IEEE International Symposium on Robot and Human Interactive Communication, 2012, pp. 771--776.
\newblock \href {https://doi.org/10.1109/ROMAN.2012.6343845} {\path{doi:10.1109/ROMAN.2012.6343845}}.

\bibitem{4415256}
A.~Edsinger, C.~C. Kemp, Human-robot interaction for cooperative manipulation: Handing objects to one another, in: RO-MAN 2007 - The 16th IEEE International Symposium on Robot and Human Interactive Communication, 2007, pp. 1167--1172.
\newblock \href {https://doi.org/10.1109/ROMAN.2007.4415256} {\path{doi:10.1109/ROMAN.2007.4415256}}.

\bibitem{Christen2023SynH2RSH}
S.~J. Christen, L.~Feng, W.~Yang, Y.-W. Chao, O.~Hilliges, J.~Song, \href{https://api.semanticscholar.org/CorpusID:265067192}{Synh2r: Synthesizing hand-object motions for learning human-to-robot handovers}, 2024 IEEE International Conference on Robotics and Automation (ICRA) (2023) 3168--3175.
\newline\urlprefix\url{https://api.semanticscholar.org/CorpusID:265067192}

\bibitem{Antanas2018SemanticAG}
L.~Antanas, P.~Moreno, M.~Neumann, R.~P. de~Figueiredo, K.~Kersting, J.~Santos-Victor, L.~D. Raedt, \href{https://api.semanticscholar.org/CorpusID:69354340}{Semantic and geometric reasoning for robotic grasping: a probabilistic logic approach}, Autonomous Robots 43 (2018) 1393 -- 1418.
\newline\urlprefix\url{https://api.semanticscholar.org/CorpusID:69354340}

\bibitem{Detry2012GeneralizingGA}
R.~Detry, C.~H. Ek, M.~Madry, J.~H. Piater, D.~Kragic, \href{https://api.semanticscholar.org/CorpusID:12062344}{Generalizing grasps across partly similar objects}, 2012 IEEE International Conference on Robotics and Automation (2012) 3791--3797.
\newline\urlprefix\url{https://api.semanticscholar.org/CorpusID:12062344}

\bibitem{Song2010LearningTC}
D.~Song, K.~Huebner, V.~Kyrki, D.~Kragic, \href{https://api.semanticscholar.org/CorpusID:10337393}{Learning task constraints for robot grasping using graphical models}, 2010 IEEE/RSJ International Conference on Intelligent Robots and Systems (2010) 1579--1585.
\newline\urlprefix\url{https://api.semanticscholar.org/CorpusID:10337393}

\bibitem{liu2020cage}
W.~Liu, A.~Daruna, S.~Chernova, Cage: Context-aware grasping engine, IEEE International Conference on Robotics and Automation (2020).
\newblock \href {http://arxiv.org/abs/1909.11142} {\path{arXiv:1909.11142}}.

\bibitem{zheng2024gaussiangrasper}
Y.~Zheng, X.~Chen, Y.~Zheng, S.~Gu, R.~Yang, B.~Jin, P.~Li, C.~Zhong, Z.~Wang, L.~Liu, C.~Yang, D.~Wang, Z.~Chen, X.~Long, M.~Wang, Gaussiangrasper: 3d language gaussian splatting for open-vocabulary robotic grasping (2024).
\newblock \href {http://arxiv.org/abs/2403.09637} {\path{arXiv:2403.09637}}.

\bibitem{Mainprice2010PlanningSA}
J.~Mainprice, E.~A. Sisbot, T.~Sim{\'e}on, R.~Alami, \href{https://api.semanticscholar.org/CorpusID:6846660}{Planning safe and legible hand-over motions for human-robot interaction}, 2010.
\newline\urlprefix\url{https://api.semanticscholar.org/CorpusID:6846660}

\bibitem{Sisbot2010SynthesizingRM}
E.~A. Sisbot, L.~F. Mar{\'i}n-Ur{\'i}as, X.~Broqu{\`e}re, D.~Sidobre, R.~Alami, \href{https://api.semanticscholar.org/CorpusID:4498998}{Synthesizing robot motions adapted to human presence}, International Journal of Social Robotics 2 (2010) 329--343.
\newline\urlprefix\url{https://api.semanticscholar.org/CorpusID:4498998}

\bibitem{Kajikawa2000TrajectoryPF}
S.~Kajikawa, E.~Ishikawa, \href{https://api.semanticscholar.org/CorpusID:60727860}{Trajectory planning for hand-over between human and robot}, Proceedings 9th IEEE International Workshop on Robot and Human Interactive Communication. IEEE RO-MAN 2000 (Cat. No.00TH8499) (2000) 281--287.
\newline\urlprefix\url{https://api.semanticscholar.org/CorpusID:60727860}

\bibitem{Ratliff2018RiemannianMP}
N.~D. Ratliff, J.~Issac, D.~Kappler, \href{https://api.semanticscholar.org/CorpusID:3707764}{Riemannian motion policies}, ArXiv abs/1801.02854 (2018).
\newline\urlprefix\url{https://api.semanticscholar.org/CorpusID:3707764}

\bibitem{nakano2010estimating}
Y.~I. Nakano, R.~Ishii, Estimating user's engagement from eye-gaze behaviors in human-agent conversations, in: Proceedings of the 15th international conference on Intelligent user interfaces, 2010, pp. 139--148.

\bibitem{lugaresi2019mediapipe}
C.~Lugaresi, J.~Tang, H.~Nash, C.~McClanahan, E.~Uboweja, M.~Hays, F.~Zhang, C.-L. Chang, M.~G. Yong, J.~Lee, W.-T. Chang, W.~Hua, M.~Georg, M.~Grundmann, Mediapipe: A framework for building perception pipelines, Third Workshop on Computer Vision for AR/VR at IEEE Computer Vision and Pattern Recognition (CVPR) (2019).
\newblock \href {http://arxiv.org/abs/1906.08172} {\path{arXiv:1906.08172}}.

\bibitem{liu2019roberta}
Y.~Liu, M.~Ott, N.~Goyal, J.~Du, M.~Joshi, D.~Chen, O.~Levy, M.~Lewis, L.~Zettlemoyer, V.~Stoyanov, Roberta: A robustly optimized bert pretraining approach, International Conference on Learning Representations (ICLR) (2019).
\newblock \href {http://arxiv.org/abs/1907.11692} {\path{arXiv:1907.11692}}.

\bibitem{wu2023brief}
T.~Wu, S.~He, J.~Liu, S.~Sun, K.~Liu, Q.-L. Han, Y.~Tang, A brief overview of chatgpt: The history, status quo and potential future development, IEEE/CAA Journal of Automatica Sinica 10~(5) (2023) 1122--1136.

\bibitem{team2023gemini}
G.~Team, R.~Anil, S.~Borgeaud, Y.~Wu, J.-B. Alayrac, J.~Yu, R.~Soricut, J.~Schalkwyk, A.~M. Dai, A.~Hauth, et~al., Gemini: a family of highly capable multimodal models, arXiv preprint arXiv:2312.11805 (2023).

\bibitem{wu2023comparative}
S.~Wu, M.~Koo, L.~Blum, A.~Black, L.~Kao, F.~Scalzo, I.~Kurtz, A comparative study of open-source large language models, gpt-4 and claude 2: Multiple-choice test taking in nephrology, NEJM AI Journal (2023).

\bibitem{spacy2}
M.~Honnibal, I.~Montani, {spaCy 2}: Natural language understanding with {B}loom embeddings, convolutional neural networks and incremental parsing, to appear (2017).

\bibitem{li2022grounded}
L.~H. Li, P.~Zhang, H.~Zhang, J.~Yang, C.~Li, Y.~Zhong, L.~Wang, L.~Yuan, L.~Zhang, J.-N. Hwang, et~al., Grounded language-image pre-training, in: Proceedings of the IEEE/CVF Conference on Computer Vision and Pattern Recognition, 2022, pp. 10965--10975.

\bibitem{lin2023wildrefer}
Z.~L. Lin, X.~Peng, P.~Cong, Y.~Hou, X.~Zhu, S.~Yang, Y.~Ma, \href{https://api.semanticscholar.org/CorpusID:258079297}{Wildrefer: 3d object localization in large-scale dynamic scenes with multi-modal visual data and natural language}, in: European Conference on Computer Vision, 2023.
\newline\urlprefix\url{https://api.semanticscholar.org/CorpusID:258079297}

\bibitem{margffoytuay2018dynamic}
E.~Margffoy-Tuay, J.~C. Pérez, E.~Botero, P.~Arbeláez, Dynamic multimodal instance segmentation guided by natural language queries, European Conference on Computer Vision (2018).
\newblock \href {http://arxiv.org/abs/1807.02257} {\path{arXiv:1807.02257}}.

\bibitem{poin_tr}
X.~Yu, Y.~Rao, Z.~Wang, Z.~Liu, J.~Lu, J.~Zhou, Pointr: Diverse point cloud completion with geometry-aware transformers, IEEE International Conference on Computer Vision (2021).
\newblock \href {http://arxiv.org/abs/2108.08839} {\path{arXiv:2108.08839}}.

\bibitem{9561877}
M.~Sundermeyer, A.~Mousavian, R.~Triebel, D.~Fox, Contact-graspnet: Efficient 6-dof grasp generation in cluttered scenes, in: 2021 IEEE International Conference on Robotics and Automation (ICRA), 2021, pp. 13438--13444.
\newblock \href {https://doi.org/10.1109/ICRA48506.2021.9561877} {\path{doi:10.1109/ICRA48506.2021.9561877}}.

\bibitem{grasp_tta}
H.~Jiang, S.~Liu, J.~Wang, X.~Wang, Hand-object contact consistency reasoning for human grasps generation, in: 2021 IEEE/CVF International Conference on Computer Vision (ICCV), 2021, pp. 11087--11096.
\newblock \href {https://doi.org/10.1109/ICCV48922.2021.01092} {\path{doi:10.1109/ICCV48922.2021.01092}}.

\bibitem{YCBDataset}
B.~Calli, A.~Singh, J.~Bruce, A.~Walsman, K.~Konolige, S.~Srinivasa, P.~Abbeel, A.~M. Dollar, \href{https://doi.org/10.1177/0278364917700714}{Yale-cmu-berkeley dataset for robotic manipulation research}, The International Journal of Robotics Research 36~(3) (2017) 261--268.
\newblock \href {http://arxiv.org/abs/https://doi.org/10.1177/0278364917700714} {\path{arXiv:https://doi.org/10.1177/0278364917700714}}, \href {https://doi.org/10.1177/0278364917700714} {\path{doi:10.1177/0278364917700714}}.
\newline\urlprefix\url{https://doi.org/10.1177/0278364917700714}

\bibitem{Zhu2021TowardHG}
T.~Zhu, R.~Wu, X.~Lin, Y.~Sun, \href{https://api.semanticscholar.org/CorpusID:244468621}{Toward human-like grasp: Dexterous grasping via semantic representation of object-hand}, 2021 IEEE/CVF International Conference on Computer Vision (ICCV) (2021) 15721--15731.
\newline\urlprefix\url{https://api.semanticscholar.org/CorpusID:244468621}

\bibitem{Lin2015RobotGP}
Y.~Lin, Y.~Sun, \href{https://api.semanticscholar.org/CorpusID:10178250}{Robot grasp planning based on demonstrated grasp strategies}, The International Journal of Robotics Research 34 (2015) 26 -- 42.
\newline\urlprefix\url{https://api.semanticscholar.org/CorpusID:10178250}

\end{thebibliography}


\begin{thebibliography}{00}

%% For numbered reference style
%% \bibitem{label}
%% Text of bibliographic item

\bibitem{lamport94}
  Leslie Lamport,
  \textit{\LaTeX: a document preparation system},
  Addison Wesley, Massachusetts,
  2nd edition,
  1994.

\end{thebibliography}

\end{document}